\documentclass[table]{ai2style/ai2}

\usepackage{microtype}
\usepackage{hyperref}
\usepackage{url}
\usepackage{booktabs, tabularx}
\usepackage{graphicx}
\usepackage{enumitem}
\usepackage{listings}

\usepackage{amssymb}
\usepackage{amsmath}
\usepackage{amsfonts}
\usepackage{multirow}
\usepackage{longtable}
\usepackage{wrapfig}
\usepackage{float}
\usepackage[most]{tcolorbox}
\usepackage{xspace}
\usepackage{nicefrac}
\usepackage[utf8]{inputenc}
\usepackage[T1]{fontenc}
\usepackage{siunitx}
\usepackage{arydshln}
\usepackage{soul}
\usepackage{adjustbox}
\usepackage{pifont}
\usepackage{caption}
\usepackage{subcaption}
\usepackage{makecell}
\usepackage{csquotes}

\definecolor{linkcolor}{RGB}{0, 0, 128}
\hypersetup{
     colorlinks   = true,
     citecolor    = linkcolor,
     linkcolor    = linkcolor,
     urlcolor     = linkcolor,
}

\setcitestyle{numbers,square,comma}

\usepackage[dvipsnames,svgnames]{xcolor}
\newcommand{\cmark}{{\color{ForestGreen}\ding{51}}}
\newcommand{\xmark}{{\color{red!70!black}\ding{55}}}

\setlist[itemize]{leftmargin=*,itemsep=0em,parsep=0.3em,topsep=0.3em}

\newcommand{\modelname}{MolmoMotion\xspace}
\newcommand{\benchmarkname}{PointMotionBench\xspace}

\graphicspath{{figures/}{.}}

\newcommand{\allenAiAff}{\raisebox{.28em}{\hspace{.02em}\scalebox{0.7}{\textbf{1}}}}
\newcommand{\uwAff}{\raisebox{.28em}{\hspace{.02em}\scalebox{0.7}{\textbf{2}}}}
\newcommand{\uncAff}{\raisebox{.28em}{\hspace{.02em}\scalebox{0.7}{\textbf{3}}}}

\newcommand{\commaAff}{\raisebox{.28em}{\hspace{.02em}\scalebox{0.7}{\textbf{,}\hspace{0.1em}}}}
\newcommand{\cofirst}{\raisebox{.28em}{\hspace{.02em}\scalebox{0.7}{\textbf{*}}}}
\newcommand{\coreContrib}{\raisebox{.28em}{\hspace{.05em}\includegraphics[height=.45em]{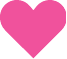}}\hspace{0.1em}}

\title{MolmoMotion\\[4pt]
  {\Large Forecasting Point Trajectories in 3D with Language Instruction}}

\authorOne[\allenAiAff\commaAff\uwAff\cofirst]{Jianing Zhang\coreContrib}
\authorOne[\allenAiAff\commaAff\uwAff\cofirst]{Chenhao Zheng\coreContrib}
\authorOne[\uwAff]{Yajun Yang\coreContrib}

\authorTwo[\allenAiAff]{Max Argus}
\authorTwo[\allenAiAff\commaAff\uwAff]{Rustin Soraki}
\authorTwo[\allenAiAff]{Winson Han}
\authorTwo[\allenAiAff]{Taira Anderson}
\authorTwo[\uwAff]{Chun-Liang Li}
\authorTwo[\allenAiAff\commaAff\uwAff]{Shuo Liu}
\authorTwo[\allenAiAff\commaAff\uwAff]{Jiafei Duan}

\authorThree[\allenAiAff\commaAff\uwAff\commaAff\uncAff]{Zhongzheng Ren\coreContrib}
\authorThree[\allenAiAff\commaAff\uwAff]{Jieyu Zhang\coreContrib}
\authorThree[\allenAiAff\commaAff\uwAff]{Ranjay Krishna\coreContrib}

\affiliation[\allenAiAff]{Allen Institute for AI}
\affiliation[\uwAff]{University of Washington}
\affiliation[\uncAff]{UNC-Chapel Hill}

\newcommand{\huggingface}{\raisebox{-1.5pt}{\includegraphics[height=1.05em]{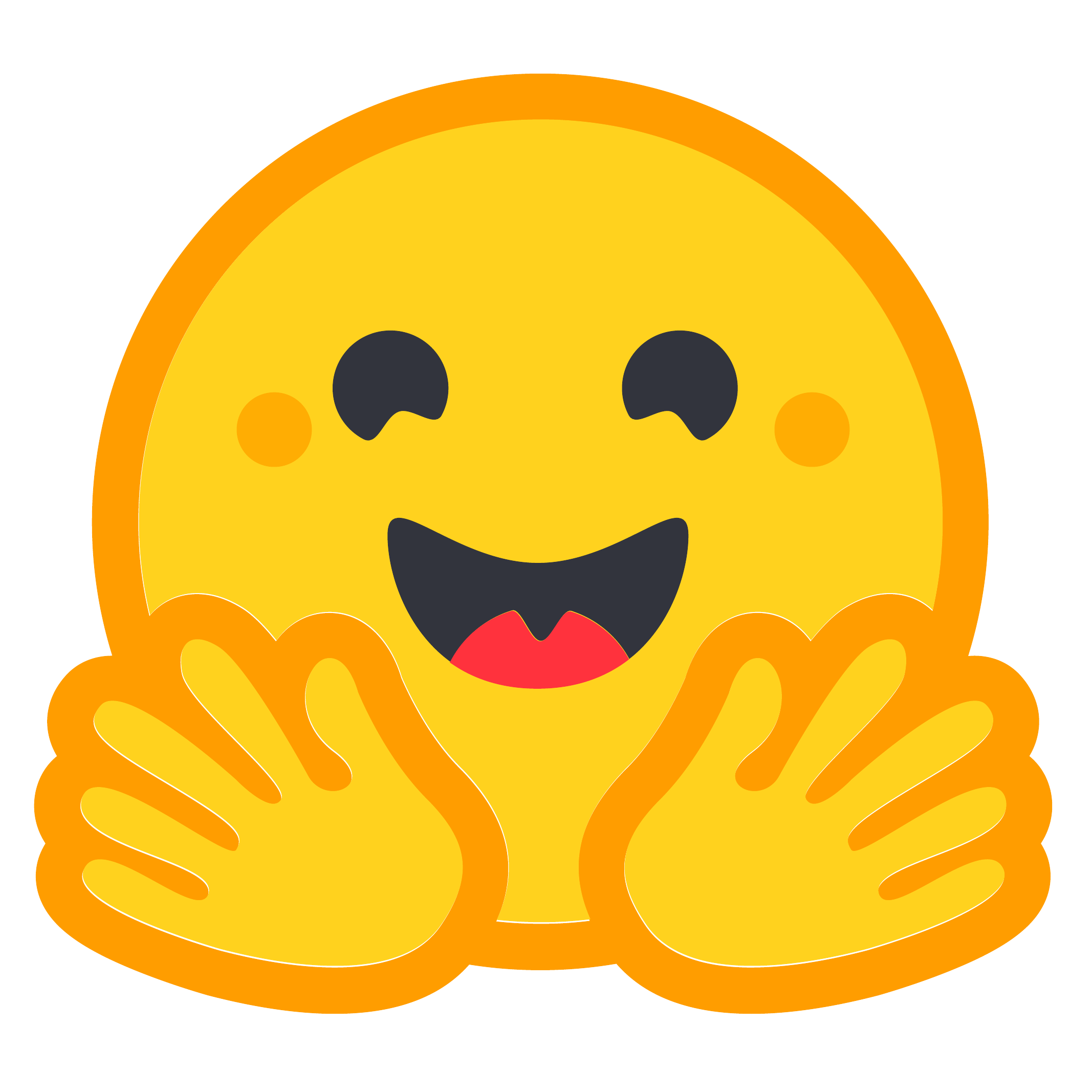}}\xspace}
\newcommand{\hfdataset}{\raisebox{-1.5pt}{\includegraphics[height=1.05em]{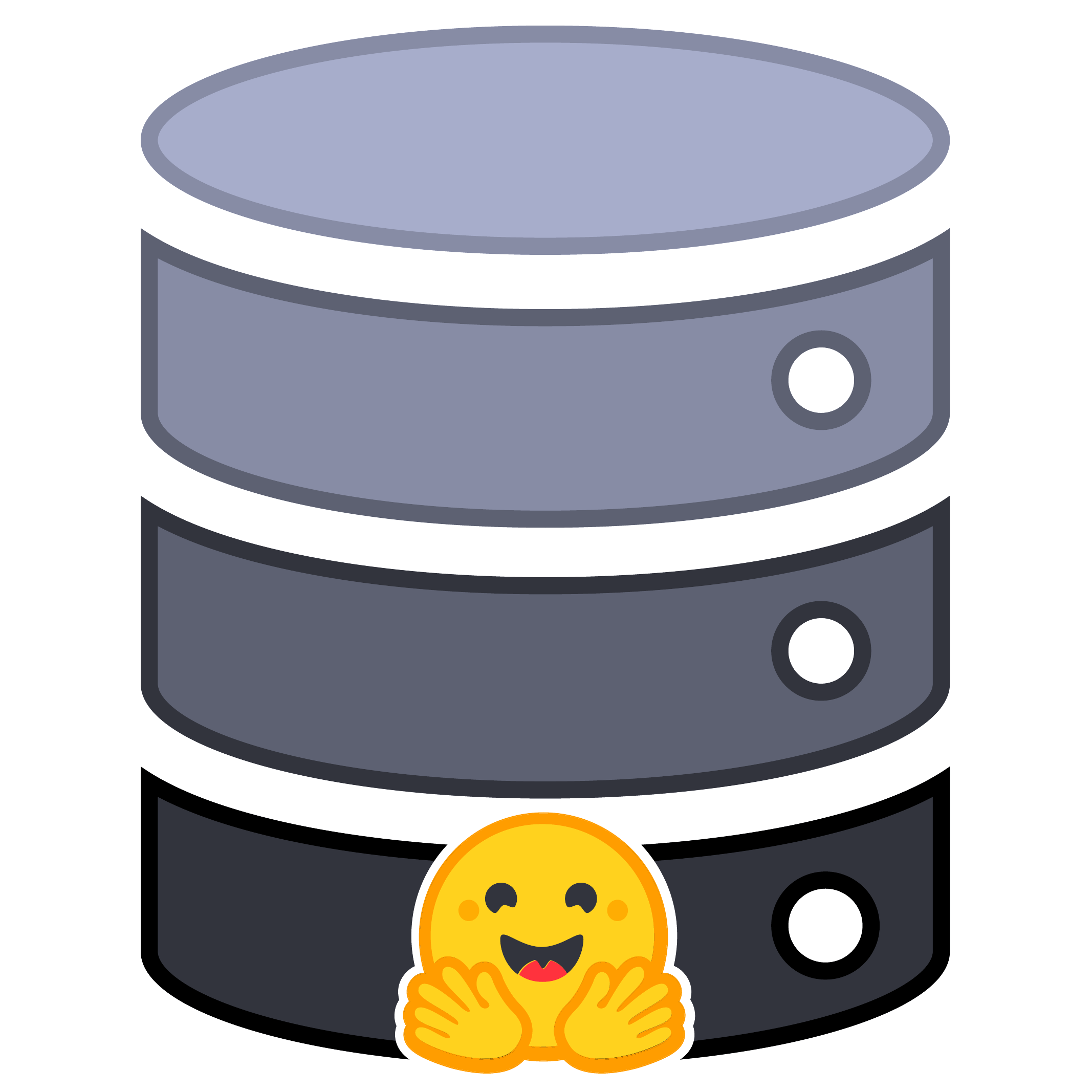}}\xspace}

\newcommand{\github}{\raisebox{-1.5pt}{\includegraphics[height=1.05em]{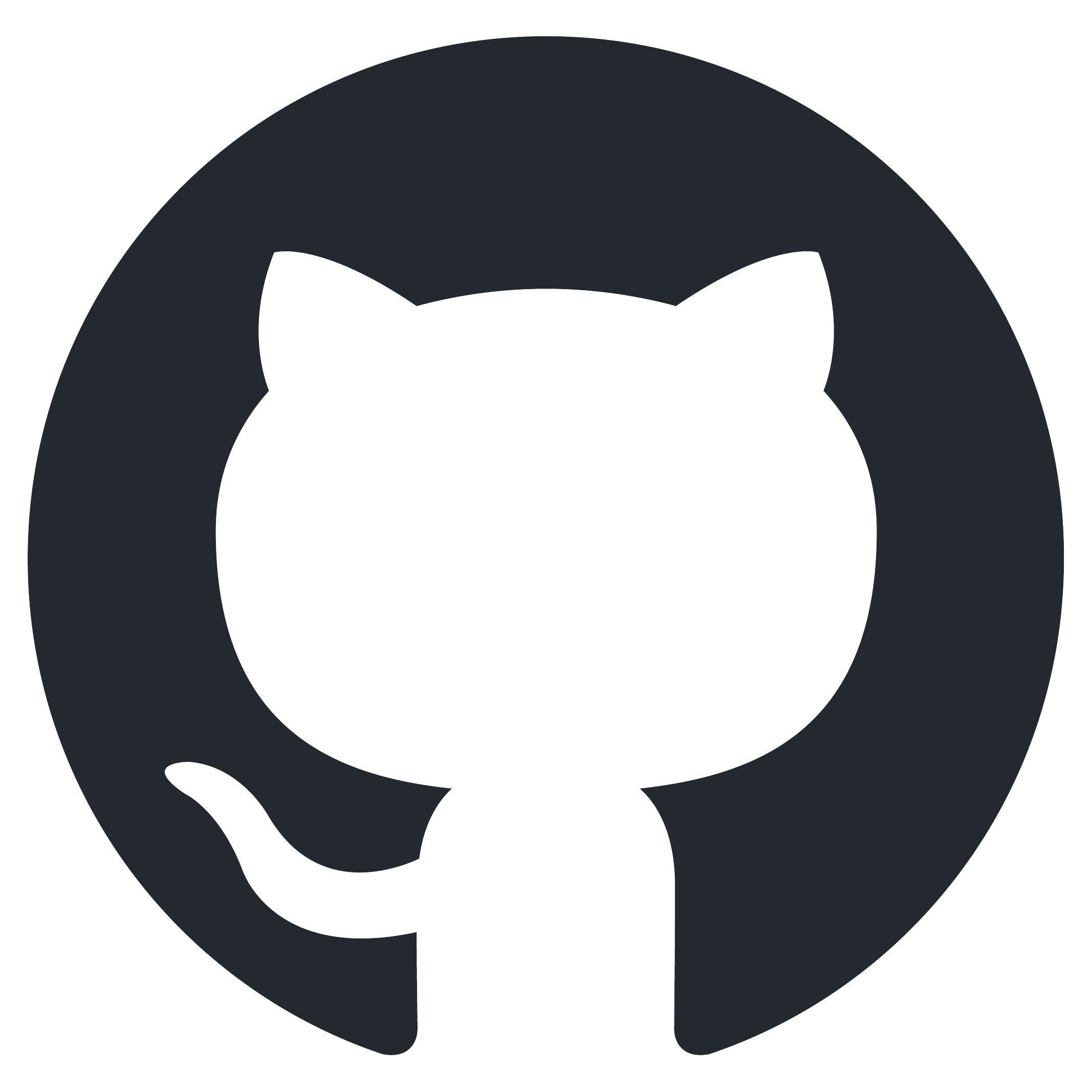}}\xspace}

\contribution[]{\raisebox{.1em}{\scalebox{0.8}{\textbf{*}}}\,Equal contribution.\quad
\raisebox{-.1em}{\includegraphics[height=.6em]{logos/core.pdf}}\,Core contributors.}

\abstract{

Motion forecasting is central to visual intelligence: agents must anticipate how objects will move in order to plan actions, reason about physical interactions, and synthesize realistic futures. We argue that 3D points in world coordinates provide a general representation that is class-agnostic, view-stable, compact, and directly useful for downstream tasks.
We formalize the task of \textit{goal-conditioned 3D point motion forecasting}: given a short visual history, a set of 3D query points on an object of interest, and a language description of the intended goal, the model predicts the future 3D trajectory of each point. We introduce a full stack to study this task at scale: (1) \textbf{MolmoMotion-1M} is a large corpus of action-described, object-grounded 3D point trajectory dataset annotated from 1.16M unconstrained videos; (2) \textbf{\benchmarkname{}} is a human-verified benchmark spanning 111 object categories and 61 motion types; and (3) \textbf{\modelname} is a general motion forecasting model that supports both autoregressive coordinate prediction and flow-matching-based trajectory generation. 
\modelname is able to accurately predicts diverse motion patterns with different language instructions, and significantly outperforms all existing motion prediction baselines on \benchmarkname{}. Finally, we show that the learned 3D motion prior transfers well to downstream applications: it improves training efficiency and generalization for robot manipulation, 
and its predicted trajectories provide effective motion guidance for generative models to synthesize videos with more realistic object motion.
}

\metadata[\quad\huggingface Models:]{
\href{https://huggingface.co/collections/allenai/molmomotion}{\texttt{\modelname}}}

\metadata[\vspace{.5em}\quad\hfdataset Data:]{
    \href{https://huggingface.co/datasets/allenai/molmo-motion-1m}{\texttt{MolmoMotion-1M}} \quad \href{https://huggingface.co/datasets/allenai/PointMotionBench}{\texttt{\benchmarkname}} 
}

\metadata[\vspace{.5em}\quad\github Code:]{
    \href{https://github.com/allenai/molmo-motion}{\texttt{https://github.com/allenai/molmo-motion}} \quad
}

\metadata[\vspace{.5em}\quad\raisebox{-1.5pt}{\includegraphics[height=1.05em]{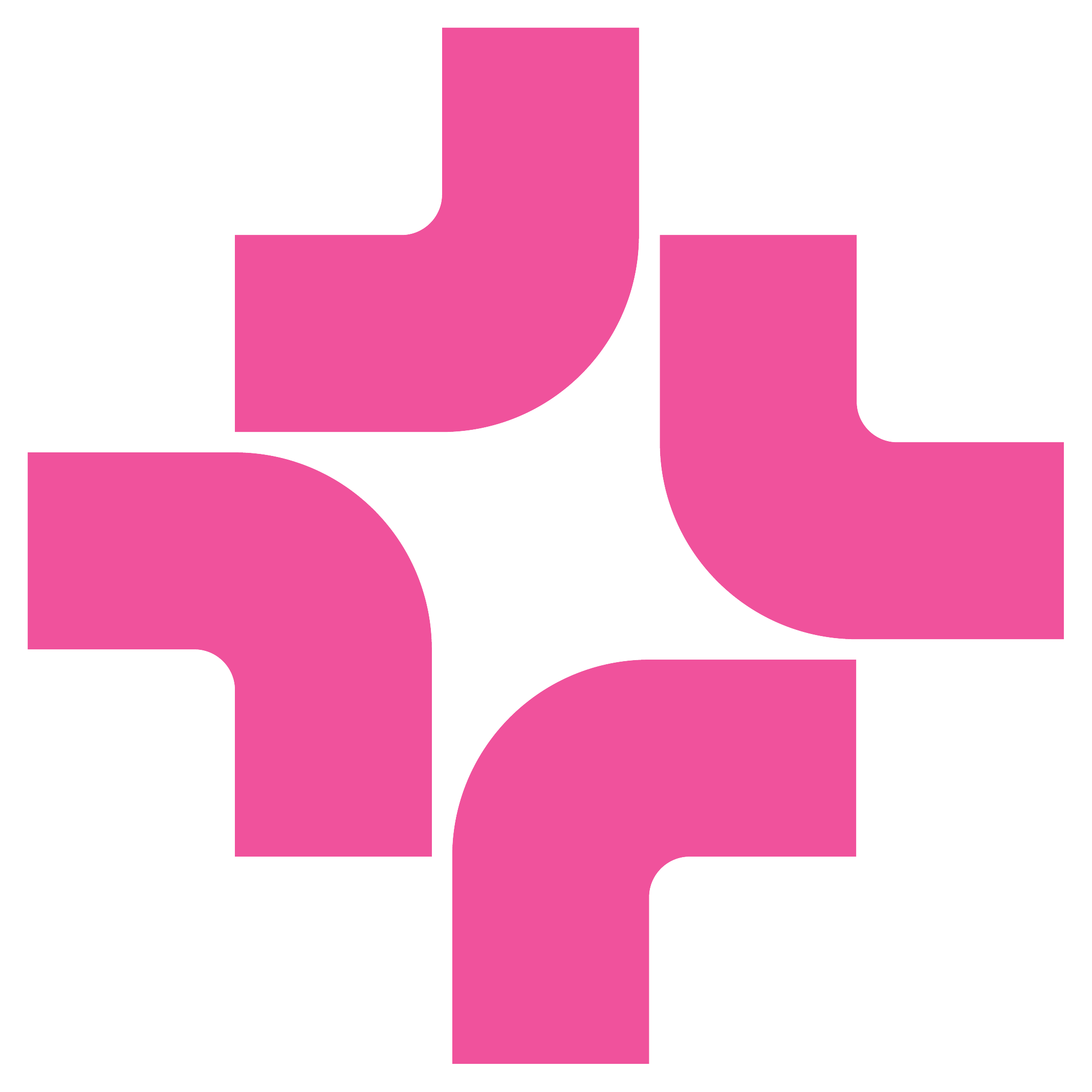}}\xspace Websites:]{\href{https://molmomotion.github.io/}{\texttt{Technical-Website}} \quad \href{https://allenai.org/blog/molmo-motion}{\texttt{Blog}}}

\begin{document}

\maketitle

\section{Introduction}
\label{sec:intro}

\begin{figure}[t]
    \centering
    \includegraphics[width=\linewidth]{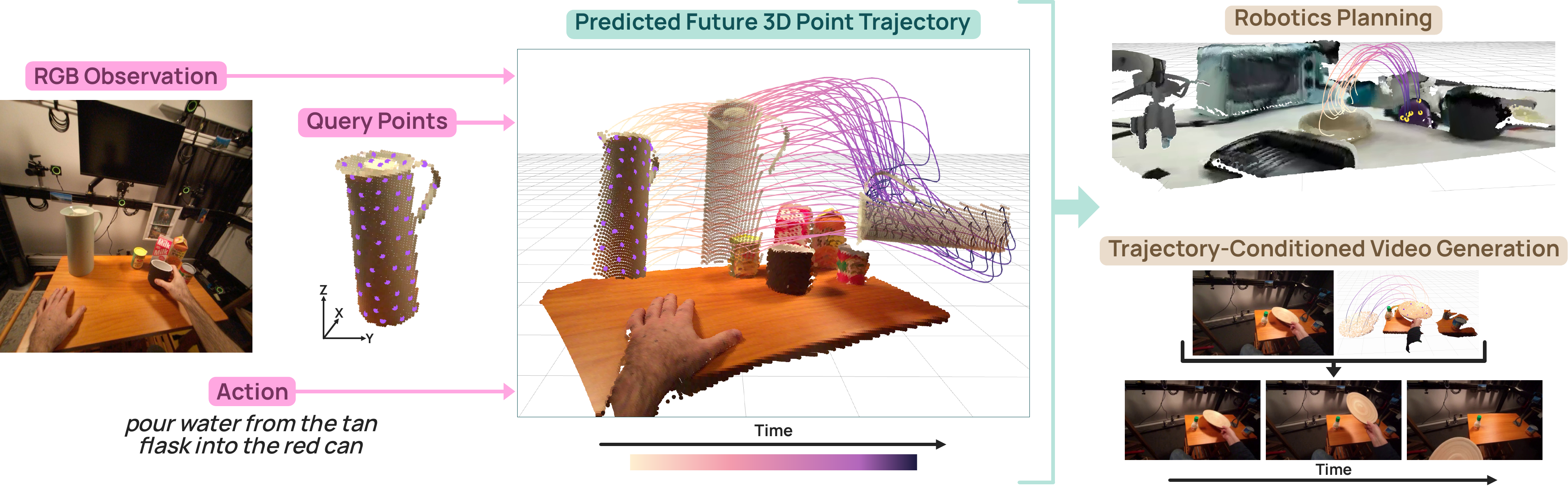}
    \caption{
        \textbf{Overview.} We introduce the task of goal-conditioned 3D point motion prediction. 
        Given initial 3D query points on an object, history RGB observations, and a language description of the future action, our model predicts the future 3D positions of all queried points in a metric world coordinate frame. 
        We show that pretraining this motion prediction task produces a transferable motion representation for downstream applications, including robotics planning and  video generation. }
    \label{fig:teaser}
\end{figure}

Psychologists like James J. Gibson have long argued that motion is core to perception, hypothesizing that motion informs an observer how they and other objects move through space, explains object occlusion and permanence, and identifies affordances~\cite{gibson2014ecological}. In the 70s, Ullman formalized motion perception as a computational problem~\cite{ullman1979interpretation}, with Lucas and Kanade providing an algorithm for estimating motion to enable tracking as optical flow~\cite{lucas1981iterative}. Although such methods have improved~\cite{raft, doersch2022tapvid, cotracker3}, they remain primarily focused on estimating motion that has \textit{already} occurred. Many real-world applications instead require \emph{forecasting} how motion will unfold. In robotics, an agent must anticipate how its actions will move objects through the scene~\cite{finn2017deepvisualforesight, track2act}. In video generation, realistic synthesis requires precise forecasting of the future motion of objects~\cite{nvidia2025cosmos, bruce2024genie}.

Building a general motion forecasting model imposes several requirements on the representation used for motion. First, the representation should be \textit{class-agnostic}: it should not depend on templates tailored to humans, hands, rigid objects, or any other fixed categories. Second, it should be \textit{view-stable}: the same underlying motion should be represented consistently across different cameras, from static surveillance footage to egocentric robot videos and moving outdoor platforms. Third, it should expose \textit{physical structure} in a form that downstream systems can use directly. 
Existing approaches of motion prediction only partially satisfy these requirements. While \textit{pixels} provide a rich rendering of possible futures~\cite{wan22, nvidia2025cosmos, bruce2024genie}, they are expensive to generate and often difficult to utilize directly in downstream applications.
Many applications instead require explicit geometric and physical quantities, such as object pose~\cite{traj2action, wen2023bundlesdf, vidbot} or particle-level dynamics~\cite{li2018particledynamics, sanchez2020graphnetworksimulators}.
\textit{Parametric 3D models} (for humans~\cite{chen2023humanmac}, hands~\cite{bi2025hrdthumanmanipulationenhanced}, or rigid objects~\cite{soraki2026objectforesightpredictingfuture3d}) are useful for such applications but remain restricted to specific categories and embodiments. 
\textit{2D point trajectories} are more category-agnostic, but image-plane coordinates entangle object motion with camera ego-motion and viewpoint change, making them difficult to disentangle from videos and transfer across domains~\cite{track2act, thakkar2026forecastingmotion}. 

We argue that \textit{object-attached 3D points in world coordinates} provide a suitable representation for general motion forecasting. A sparse set of surface points moving through 3D space can describe the motion of rigid, articulated, or deformable objects without assuming a category-specific template. Because 3D points share a world frame, the same physical motion can remain stable across cameras, viewpoints, and capture settings. Additionally, many task-relevant quantities can be expressed as the motion of 3D points, from the pose change of a robot gripper~\cite{vidbot} to particle trajectories in physical simulation~\cite{li2018particledynamics, sanchez2020graphnetworksimulators}. 
Finally, by focusing only on object-attached points, the prediction target remains compact while still capturing the motion relevant for physical interaction.

We formalize this idea as the task of \textit{goal-conditioned 3D point motion prediction} (Fig.~\ref{fig:teaser}). Given a short history of visual observations, a set of initial 3D query points on an object of interest, and a language description of the intended goal, the model predicts the future 3D trajectory of each query point over time. The language instruction disambiguates among plausible futures and reduces the search space of possible future states.

We present the full stack needed to study this task at scale: \textbf{a scalable data curation pipeline}, \textbf{a novel prediction model}, and \textbf{a human-verified benchmark}. The first challenge is 3D motion supervision data. Existing video datasets with 3D capture are small and domain-limited~\cite{hot3d, koppula2024tapvid3d, liu2022hoi4d}, while internet videos provide scale and diversity but lack 3D annotations. We therefore develop an automatic annotation pipeline for robustly extracting object-grounded 3D point trajectories from unconstrained videos. Applying this pipeline to roughly 1.16M video clips yields MolmoMotion-1M, the largest corpus of action-described, object-grounded 3D point trajectories. We additionally introduce \benchmarkname{}, a benchmark for 3D motion forecasting spanning 111 object categories and 61 motion types. The benchmark uses ground-truth 3D capture where available and human-verified 3D tracks otherwise, providing a reliable testbed for evaluating motion prediction.

We introduce \modelname, a general motion prediction model pretrained on the MolmoMotion-1M dataset. We train two complementary classes of motion prediction models. The first predicts future trajectories autoregressively as coordinate sequences; the second predicts future motion with flow matching as a continuous trajectory distribution. 
Experiments show \modelname accurately forecast diverse motion types and generalize across a wide range of scenes and instructions, and significantly outperforms all existing motion prediction methods in \benchmarkname{}.

Finally, we validate 3D point forecasting as a useful task: \modelname transfers effectively to downstream applications.
Because object trajectories in 3D world space are largely embodiment-agnostic~\cite{track2act, dharmarajan2025dream2flow}, the motion prior learned from internet-scale human video can transfer naturally to robotics planning. We show that this prior improves sample efficiency, closed-loop success, and generalization to unseen objects and scenes in the MolmoSpaces pick-and-place task~\cite{molmospaces}, and adapts efficiently to real-world robot videos from DROID~\cite{droid}.
\modelname can also guide video generation. When its predicted trajectories are used to condition a lightweight 3D-track-conditioned image-to-video model~\cite{gu2025diffusionasshader}, the generated videos exhibit more realistic motion than those produced directly by much larger image-to-video models~\cite{wan22}, while also improving multiple quantitative video-quality metrics.

\section{\modelname}

We formulate general motion prediction as the task of predicting future 3D trajectories of points attached to an object of interest, conditioned on an action description.
In this section, we formulate the problem (Sec.~\ref{subsec: problem_formulation})
and introduce our model architecture (Sec.~\ref{subsec:model}).

\subsection{Problem formulation}
\label{subsec: problem_formulation}
We define the model input as follows. At a reference time $t_0$, the model is given $N$ user-specified 2D query points on an object of interest in the image,
$\{\mathbf{q}_{t_0}^{n} \in \mathbb{R}^2\}_{n=1}^{N}$.
It is also given the corresponding initial 3D positions
$\{\mathbf{p}_{t_0}^{n} \in \mathbb{R}^3\}_{n=1}^{N}$,
expressed in the camera coordinate frame at $t_0$. In practice, these 3D positions can be obtained by lifting the 2D query points using estimated or measured depth together with known camera intrinsics. 
The model also receives a short history of RGB observations, $I_{t_s:t_0}=\{I_{t_s},\ldots,I_{t_0}\}$, and a language description $a$ of the intended action.
The goal is to predict the future 3D positions of all query points,
over a horizon of $T$ steps: 
$\{\{\hat{\mathbf{p}}_{t}^{n}\in\mathbb{R}^3\}_{t=t_0+1}^{t_0+T}\}_{n=1}^{N}$.
All future coordinates are expressed in a world coordinate frame anchored at the camera at time $t_0$. This choice makes the prediction independent of future camera motion.

\begin{figure}[t]
    \centering
    \includegraphics[width=0.93\linewidth]{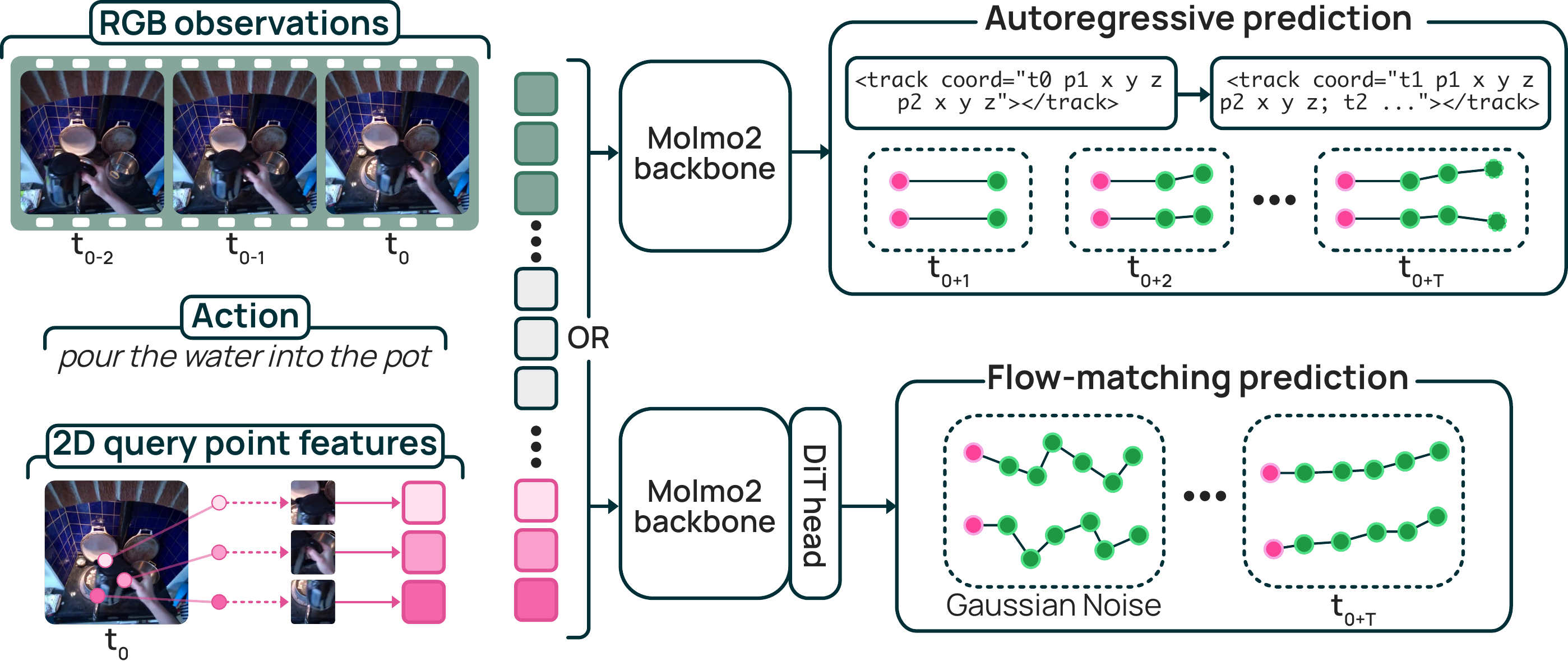}
    \caption{\textbf{\modelname architecture.}
    The shared input to Molmo2~\cite{clark2026molmo2} backbone consists of image tokens of RGB observations, text tokens of action description, and 2D query point feature tokens sampled from Molmo2 vision encoder. 
    The autoregressive variant encodes the initial 3D query coordinates and decodes future trajectories as quantized coordinate text, while the flow-matching variant represents them directly in continuous 3D coordinate space. }
    \label{fig:architecture}
\end{figure}

\subsection{\modelname architecture}
\label{subsec:model}

We implement two classes of trajectory predictors: one with an autoregressive objective and the other with a flow-matching objective. They share the same input encoding of the RGB observation, action description, and 2D query-point visual features.
They differ in how the initial 3D query coordinates are encoded and how the future 3D trajectories are decoded (Fig.~\ref{fig:architecture}).
We study both objectives because they offer complementary modeling biases: autoregressive decoding conditions each prediction on the previously generated trajectory, which encourages smooth temporal evolution, while flow-matching models a distribution over future trajectories and is better suited for capturing motion uncertainty~\cite{chi2024diffusionpolicyvisuomotorpolicy}.

\noindent\textbf{Input encoding.}
Predicting object motion from a language instruction first requires grounding the relevant object in the image and understanding the instruction. We therefore use Molmo2~\cite{clark2026molmo2} as the vision-language backbone for input processing, leveraging its strong object-grounding capability. 
Given the RGB observation history $I_{t_s:t_0}$, the Molmo2 vision encoder produces image tokens $\mathcal{T}_{\mathrm{img}}$. The action description $a$ is tokenized into language tokens $\mathcal{T}_{\mathrm{text}}$. To condition on the query points, we additionally insert one visual point token for each 2D query coordinate. Let $F_{t_0}$ be the anchor-frame feature map produced by the vision encoder. 
For each query point $\mathbf{q}_{t_0}^{n}$, we bilinearly sample the anchor-frame feature map $F_{t_0}$ at its 2D location to obtain a point feature $\mathbf{e}_{\mathrm{pt}}^{n}$.
These features form the point-token sequence
$\mathcal{T}_{\mathrm{pt}}=\{\mathbf{e}_{\mathrm{pt}}^{1},\ldots,\mathbf{e}_{\mathrm{pt}}^{N}\}$.
We concatenate the image, text, and point tokens as
$\mathcal{C}=[\mathcal{T}_{\mathrm{img}},\mathcal{T}_{\mathrm{text}},\mathcal{T}_{\mathrm{pt}}]$
and process them with the Molmo2 language model component.

For both prediction variants, we represent 3D coordinates relative to the first query point at $t_0$. Let $\mathbf{p}_{\mathrm{anc}}=\mathbf{p}_{t_0}^{1}$ denote this anchor point. For each point $n$ and time $t$, our models represent 3d point coordinates as
$\boldsymbol{\delta}_{t}^{n}
=
\mathbf{p}_{t}^{n}-\mathbf{p}_{\mathrm{anc}}$.
All 3D point coordinates are in metric scale, in units of \textit{meter}.

\noindent\textbf{Autoregressive objective.}
The autoregressive variant follows Molmo2's coordinate representation, encoding both the initial 3D query coordinates and the future trajectory as structured text. Each anchor-relative coordinate is discretized into \textit{millimeter} bins
($\bar{\boldsymbol{\delta}}_{t}^{n}
=
\mathrm{round}\!\left(
1000\,\boldsymbol{\delta}_{t}^{n}
\right)$)
and serialized as timestamped point-coordinate tuples. Fig.~\ref{fig:architecture} shows an example of the coordinate encoding format.
The input prompt contains the visual-language conditioning $\mathcal{C}$ together with the serialized initial query coordinates $\{\bar{\boldsymbol{\delta}}_{t_0}^{n}\}_{n=1}^{N}$. The output is the serialized future trajectory $y_{1:L}$, generated in temporal order. An example of text pieces encoded and decoded can be found at Fig.~\ref{fig:architecture}.
We train the autoregressive model with the standard next-token objective.
At inference time, the model decodes the trajectory string autoregressively and the coordinates are parsed back into $\hat{\mathbf{P}}_{t_0+1:t_0+T}$. Since coordinates are emitted in temporal order, each future timestamp is conditioned on all earlier generated coordinates, giving the model a direct mechanism for modeling temporal dependence and smooth rollouts.

\noindent\textbf{Flow-matching objective.}
The flow-matching variant predicts future trajectories in continuous 3D coordinate space. Following a MolmoBot-style design~\cite{deshpande2026molmobot}, we use a DiT decoder~\cite{peebles2023dit} conditioned on Molmo2 features from all layers, with lightweight MLPs for coordinate encoding and decoding. We concatenate the clean initial 3D query coordinates with a noised version of the future coordinates and project them into point tokens. We use RoPE along both the point and time axes to distinguish point identity and timestamp, analogous to the Multimodal RoPE used in Qwen2.5-VL~\cite{su2023roformerenhancedtransformerrotary,bai2025qwen25vl}.

The decoder is trained with the standard flow-matching objective~\cite{lipman2023flowmatching}. Specifically, we sample Gaussian noise with the same shape as the future trajectory, linearly interpolate between the noise and the ground-truth trajectory, and train the decoder to predict the corresponding velocity field from the noised trajectory to the clean trajectory. At inference time, the model starts from Gaussian noise and integrates the learned velocity field with 10 Euler steps to obtain the predicted future trajectory.

\begin{figure}[t]
    \centering
    \includegraphics[width=0.93\linewidth]{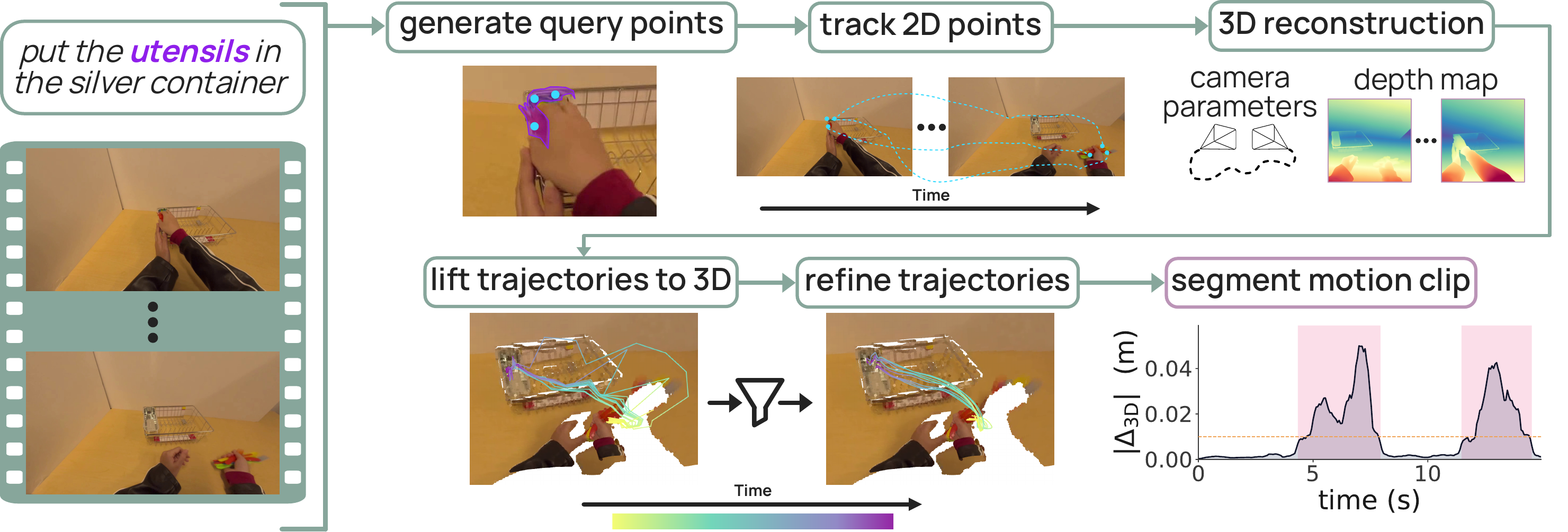}
    \caption{\textbf{Overview of data annotation pipeline.}  Given a video of an action event and its description, we first ground the moving object and sample query points on it. We then track dense 2D points on the object, lift these tracks into a shared metric 3D frame, and use object-level spatial and temporal consistency priors to filter unreliable trajectories. Finally, we clip the video around intervals where the grounded object undergoes meaningful motion.}
    \label{fig:pipeline}
\end{figure}

\section{MolmoMotion-1M and \benchmarkname{}}

Existing 3D point-track datasets are small, domain-limited, and often lack object grounding or language annotations. We therefore build an automatic annotation pipeline that extracts object-grounded 3D point trajectories from unconstrained videos. Applying it to 1.16M public videos yields \textbf{MolmoMotion-1M}, the largest action-described, object-grounded 3D point trajectory dataset. We also introduce \textbf{\benchmarkname{}}, a held-out benchmark with verified 3D point trajectories for object-centric 3D motion forecasting.

\subsection{MolmoMotion-1M Annotation}
\label{subsec:motion_annotation}

We start with public video datasets that provide action descriptions or task captions~\cite{egodex, hdepic, xperience, stereo4d}. Our goal is to localize the described object, select query points, and track them in 3D world coordinates. The pipeline consists of five stages, shown in Fig.~\ref{fig:pipeline}: semantic object grounding, temporal point correspondence, metric 3D lifting, trajectory-level filtering, and video-level clipping.

\noindent\textbf{Semantic object grounding.}
Given an action description $a$, we use an LLM~\cite{yang2025qwen3} to extract the manipulated or moving entity as a short object phrase. Instead of prompting SAM3~\cite{sam3} to directly produce an object mask from this phrase, we first use MolmoPoint~\cite{clark2026molmopoint} to localize the entity as a 2D point in the reference frame $t_0$. We find MolmoPoint more reliable for vague phrases like ``an object on top of the table,'' because the prompt can specify the object by its action, e.g., the object being moved, rather than by appearance alone. We then use the point prompt with SAM3 to obtain the object mask $M_{t_0}$, and sample $N$ query points $\{\mathbf{q}_{t_0}^{n}\}_{n=1}^{N}$ inside $M_{t_0}$ using K-means cluster centers.

\noindent\textbf{2D point tracking and metric 3D lifting.}
We next establish point correspondence across frames by tracking the query points through the video. We run AllTracker~\cite{alltracker} to obtain temporally persistent 2D tracks  $\{\mathbf{q}_{t}^{n}\}_{t=1}^{L}$ and visibility masks $\{m_t^{n}\}_{t=1}^{L}$.
To lift these tracks into 3D, we run ViPE~\cite{vipe} on the monocular video to estimate per-frame metric depth and camera geometry. We empirically find that this paradigm produces more accurate 3D trajectories than current end-to-end 3D point trackers such as SpatialTrackerV2~\cite{spatialtrackerv2}. ViPE also provides metric scale output, allowing the resulting trajectories to be expressed in physical 3D units rather than a arbitrary relative scale.  Using the estimated depth, intrinsics, and camera poses, we back-project each visible 2D track point into a world frame anchored at the first-frame camera, producing metric 3D tracks $\{\tilde{\mathbf{p}}_{t}^{n} \in \mathbb{R}^3\}_{t=1}^{L}$.

\noindent\textbf{Trajectory-level filtering and smoothing.}
We empirically find that some lifted trajectories can be corrupted by 2D tracking drift, depth noise, and camera estimation errors. 
We therefore both filter outlier tracks and smooth remaining tracks using the object-level prior that sampled points should move coherently as parts of the same physical entity.
We remove tracks with consistently low trust using a MAD-based outlier criterion~\cite{leys2013madoutlier}.
For retained tracks, we follow the smoothing algorithm in Stereo4D~\cite{stereo4d} to smooth their depth value along each camera ray. More details can be found in Appendix~\ref{appsec:filtering}.
This removes high-frequency depth jitter in annotated 3D tracks.

\noindent\textbf{Video-level clipping.}
Event-level video clips often contain long static intervals before or after the described motion, which provide little supervision for learning dynamics. We therefore re-clip videos around intervals where the grounded object actually moves.
Given the filtered 3D trajectories, we compute a per-frame object motion score $s_t$ as the median 3D displacement of valid object points:
$
s_t
=
\mathrm{median}_{n}
\left\|
\mathbf{p}_{t}^{n}
-
\mathbf{p}_{t-1}^{n}
\right\|_2 .
$
We threshold $s_t$ to obtain contiguous segments with non-trivial object motion. This step also automatically removes static videos.

\begin{figure}[t]
\centering
\begin{subfigure}{0.32\linewidth}
    \includegraphics[width=\linewidth]{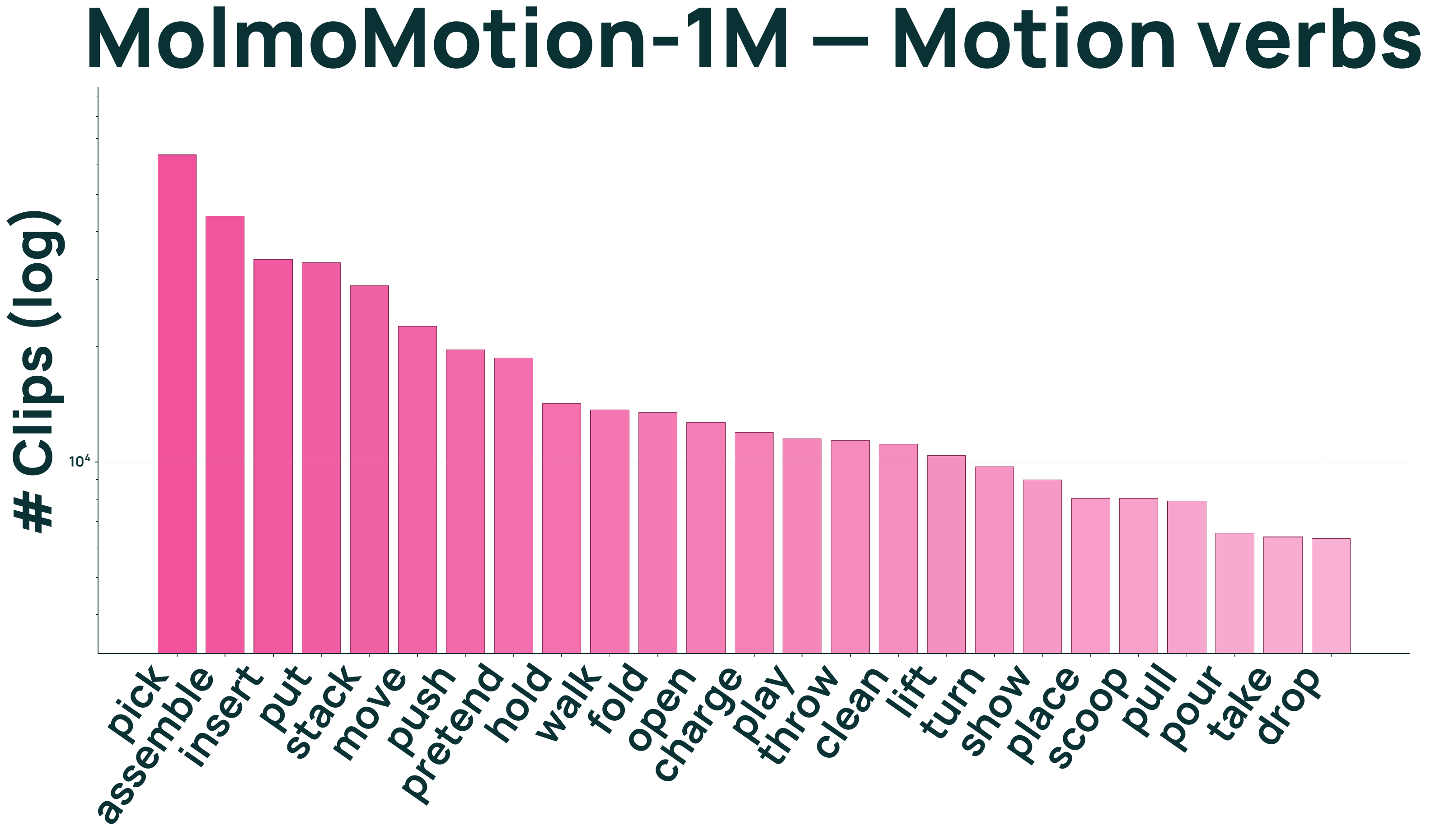}
    \caption{MolmoMotion-1M (pretrain).}
\end{subfigure}\hfill
\begin{subfigure}{0.32\linewidth}
    \includegraphics[width=\linewidth]{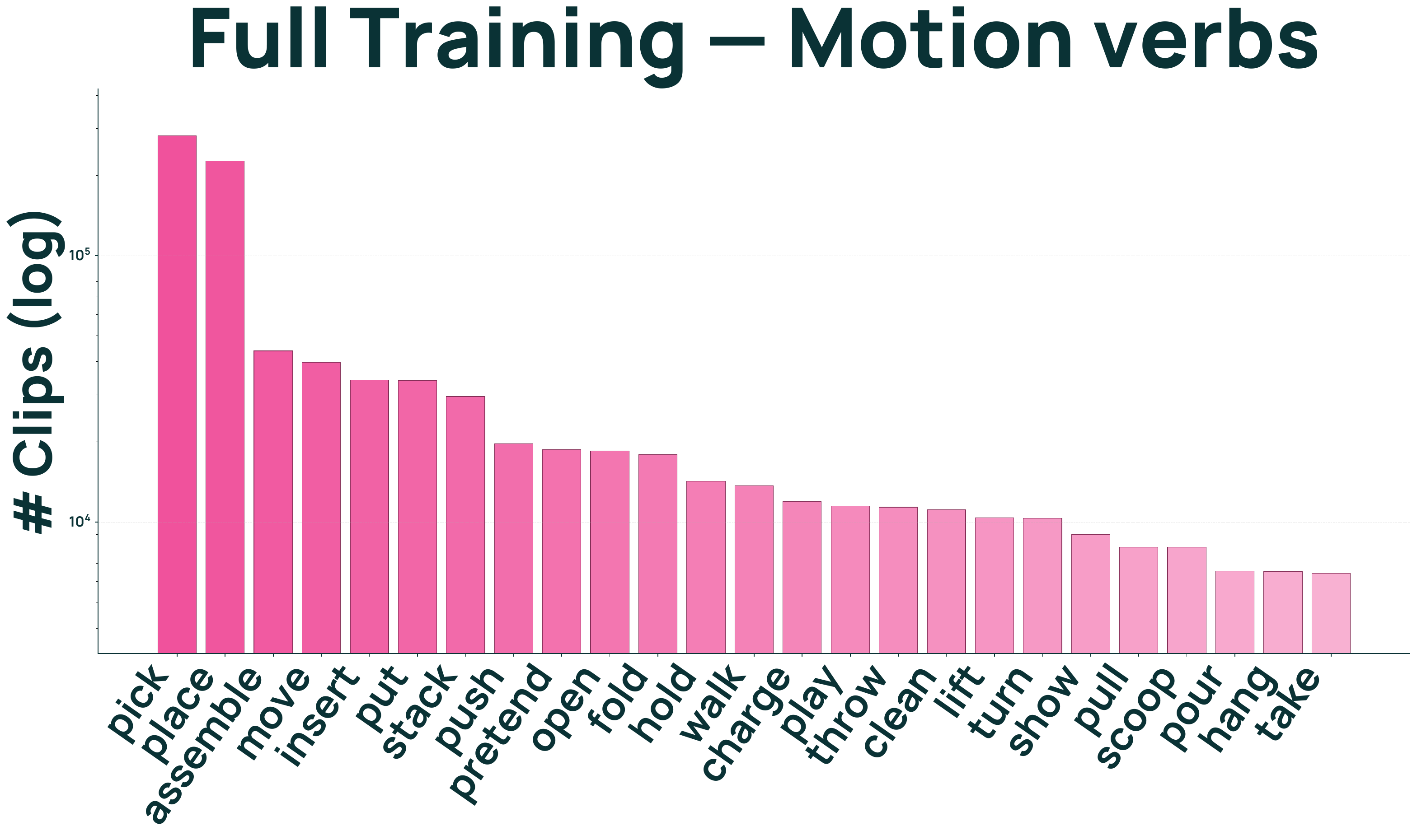}
    \caption{Pretrain + downstream.}
\end{subfigure}\hfill
\begin{subfigure}{0.32\linewidth}
    \includegraphics[width=\linewidth]{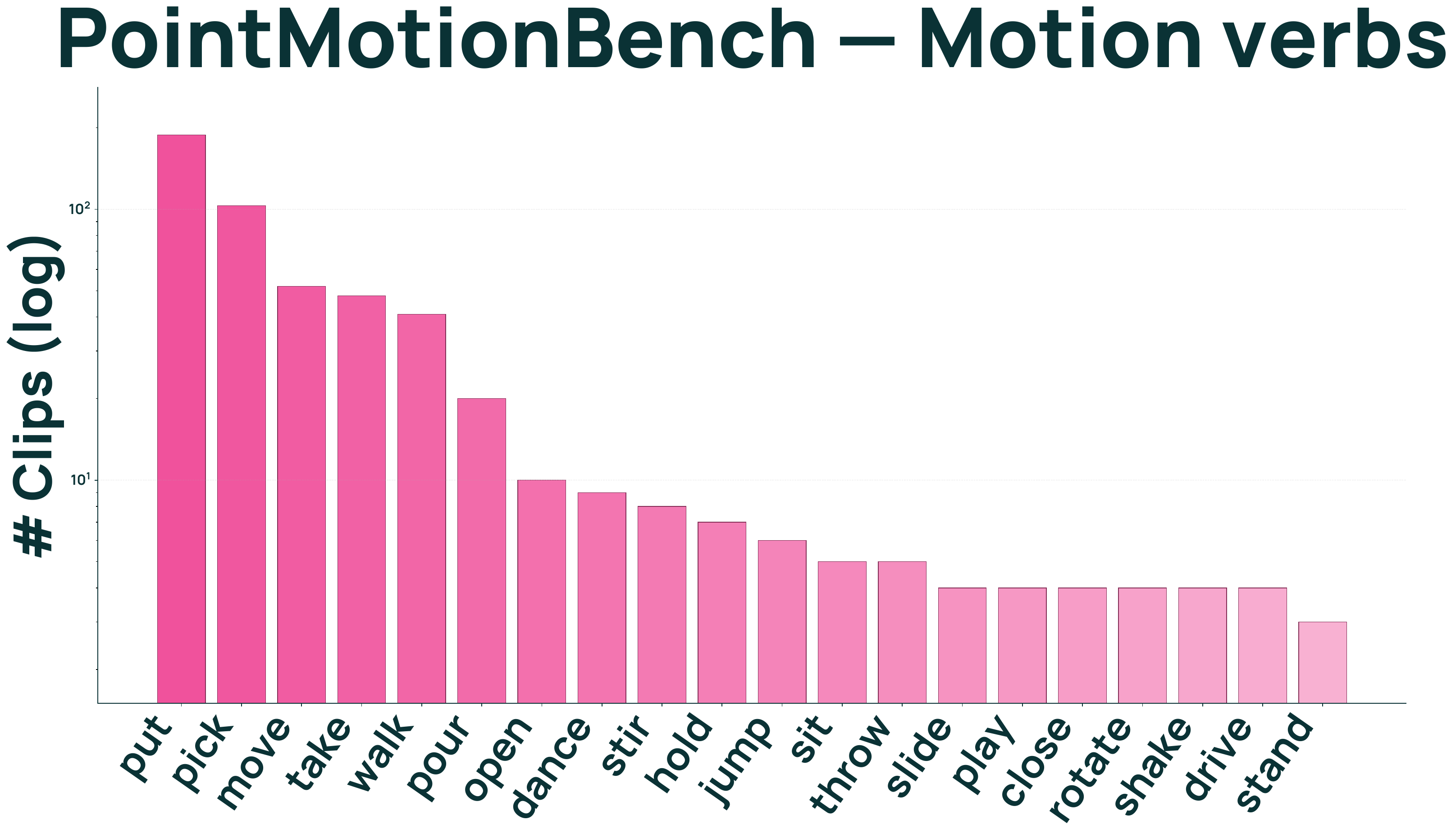}
    \caption{\benchmarkname{}.}
\end{subfigure}
\caption{\textbf{Action-verb diversity.} Distribution of the most frequent motion verbs across the pretraining corpus (MolmoMotion-1M), the full training corpus that additionally includes the MolmoSpaces and DROID downstream-finetune data, and the \benchmarkname{} evaluation set. Bars are log-scaled.}
\label{fig:diversity-verbs}
\end{figure}

\begin{figure}[t]
\centering
\begin{subfigure}{0.32\linewidth}
    \includegraphics[width=\linewidth]{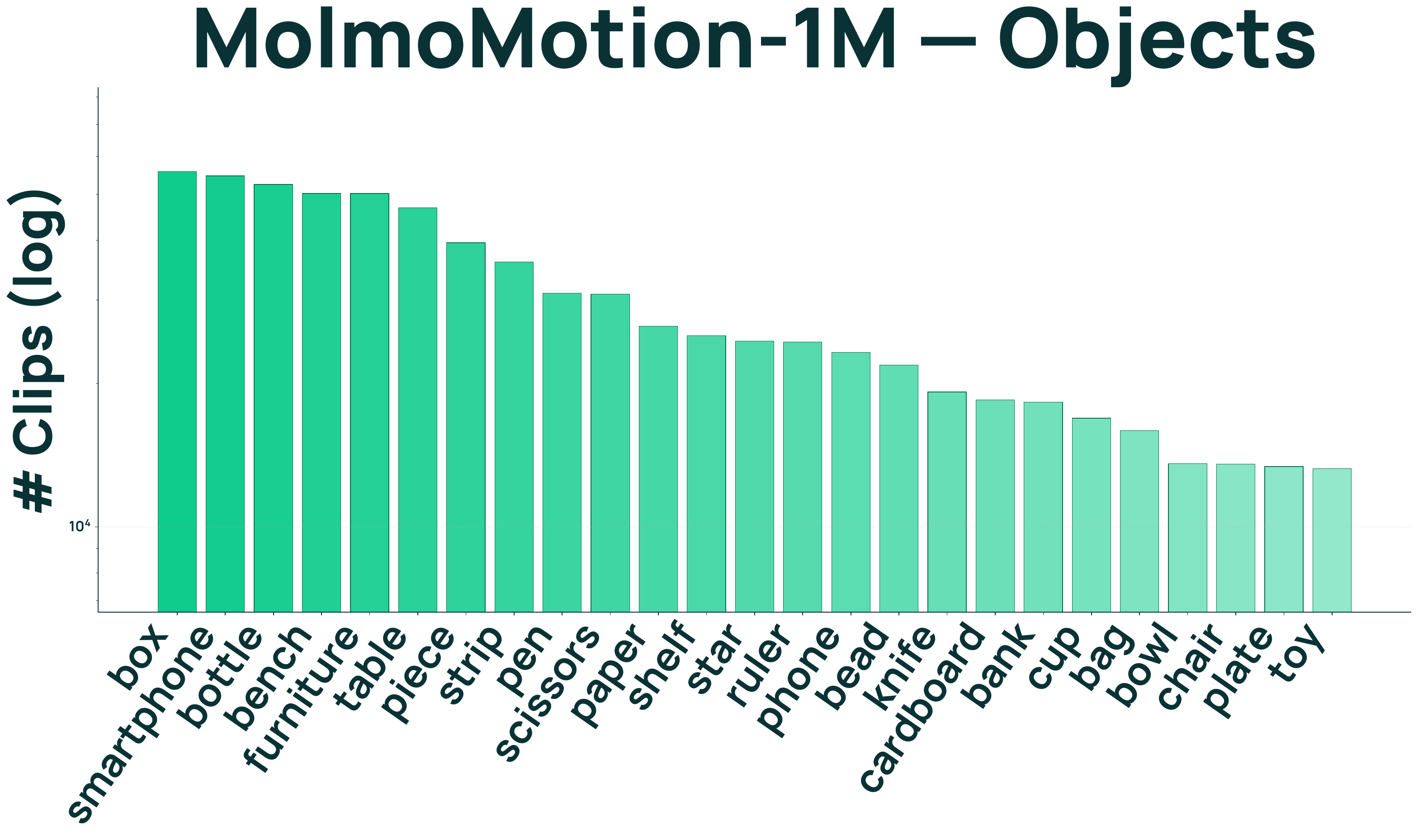}
    \caption{MolmoMotion-1M (pretrain).}
\end{subfigure}\hfill
\begin{subfigure}{0.32\linewidth}
    \includegraphics[width=\linewidth]{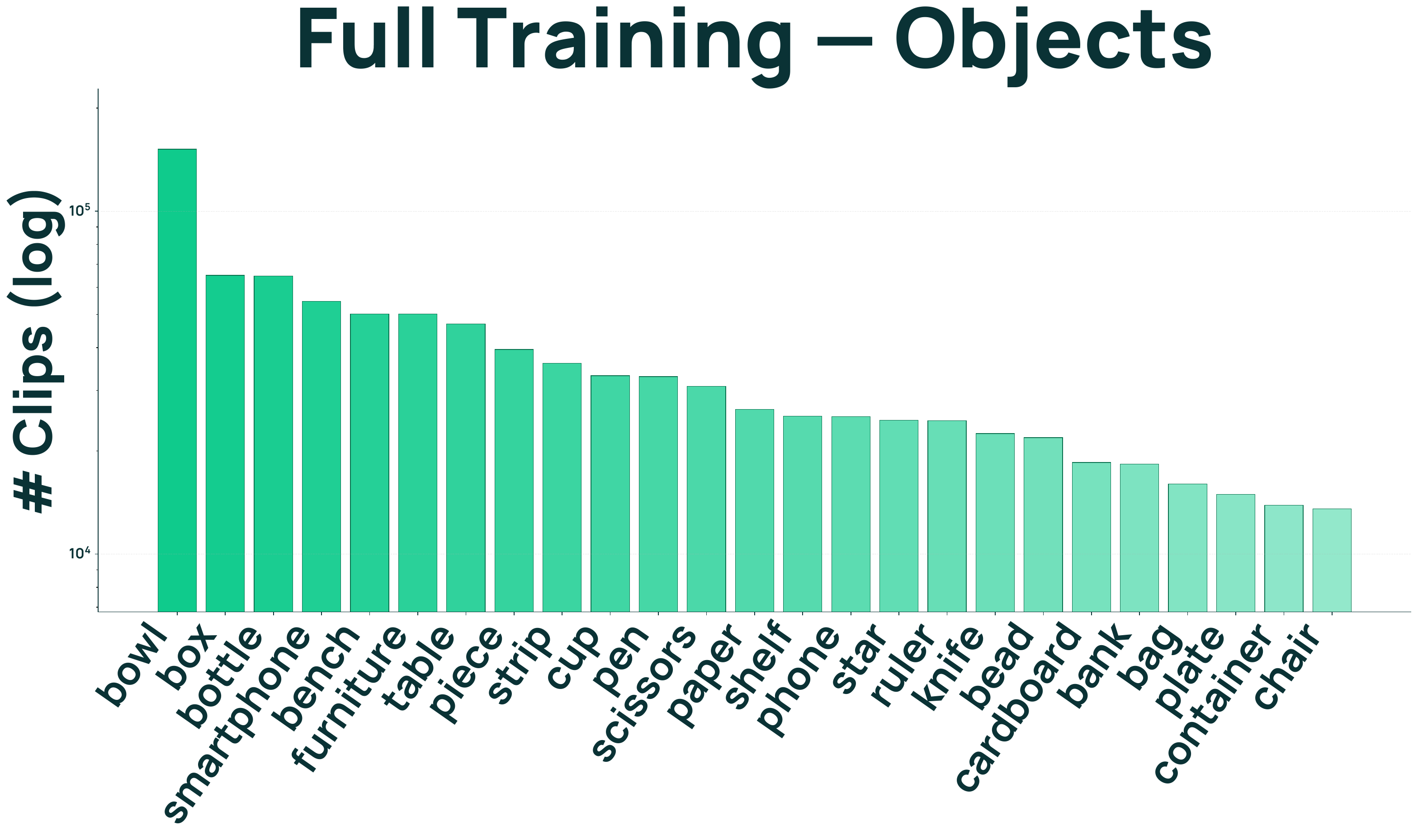}
    \caption{Pretrain + downstream.}
\end{subfigure}\hfill
\begin{subfigure}{0.32\linewidth}
    \includegraphics[width=\linewidth]{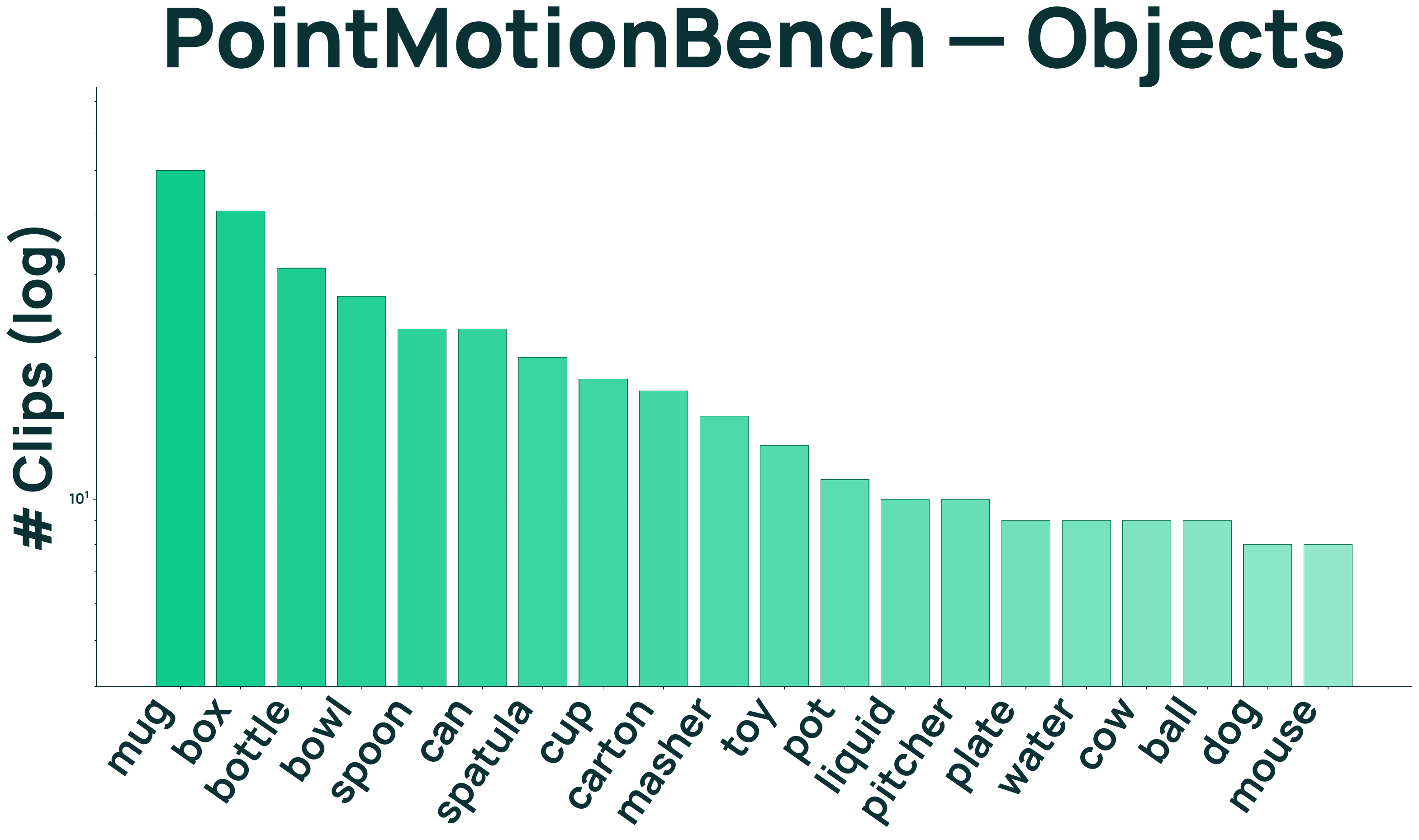}
    \caption{\benchmarkname{}.}
\end{subfigure}
\caption{\textbf{Manipulated-object diversity.} Distribution of the most frequent manipulated objects per cohort. Generic placeholder tokens (\textit{object}, \textit{hand}, \textit{person}) are excluded. Bars are log-scaled.}
\label{fig:diversity-objects}
\end{figure}

\noindent\textbf{Pretraining corpus statistics.}
Our videos span human manipulation, hand-object interaction, and in-the-wild scenarios. The largest portion comes from human-object manipulation datasets: EgoDex~\cite{egodex}, HD-EPIC~\cite{hdepic}, and Xperience-10M~\cite{xperience} together contribute the bulk of egocentric and third-person manipulation clips; we additionally include YT-VIS~\cite{ytvis} and Stereo4D~\cite{stereo4d} clips, which broaden coverage to outdoor scenes and deformable objects like animals. 

After filtering and segmentation, the pipeline produces approximately $1$M clips with motion (per-corpus counts in Tab.~\ref{tab:appendix-sources}). The corpus spans $736$ unique action verbs and $5{,}692$ unique manipulated objects. Clips are short by construction: median clip length is $0.8$--$1.1$\,s on the manipulation corpora and $1.7$\,s on Stereo4D, where third-person walking subjects yield longer motion windows. Median per-clip 3D displacement ranges from $7$--$9$\,cm on the manipulation corpora to $51$\,cm on Stereo4D, reflecting the difference between tabletop manipulation and walking subjects. After filtering, each clip retains a median of $88$ query points (range $60$--$100$). Fig.~\ref{fig:diversity-verbs} and Fig.~\ref{fig:diversity-objects} show the per-cohort distributions of motion verbs and manipulated objects across the pretraining corpus (MolmoMotion-1M), and the full training corpus that additionally includes the MolmoSpaces and DROID data used for downstream robot finetuning.

\subsection{\benchmarkname{}}
\label{subsec:data_and_benchmark}
We also introduce \textbf{\benchmarkname{}}, a held-out benchmark for evaluating object-centric 3D motion forecasting. Unlike the pretraining corpus, which starts from videos with action descriptions, \benchmarkname{} repurposes datasets with ground-truth 3D capture to ensure annotation accuracy. For HOT3D~\cite{hot3d} and WorldTrack~\cite{feng2025st4rtrack}, we extract 3D point trajectories directly from the provided 3D object mesh or points, ground the foreground points to objects, and annotate action descriptions, with all annotations verified by humans (details in Appendix~\ref{subsec:appendix_3dworldbench}). Since HOT3D and WorldTrack mainly cover indoor manipulation and egocentric hand-object interaction, we further include DAVIS~\cite{davis} to cover outdoor dynamic scenes; for this split, we run our annotation pipeline and manually verify the correctness of each resulting trajectory.
In total, the benchmark contains $\mathbf{742}$ clips spanning $\mathbf{111}$ object categories and $\mathbf{61}$ action/motion types. Fig.~\ref{fig:diversity-verbs}(c) and Fig.~\ref{fig:diversity-objects}(c) show the per-cohort distributions of motion verbs and manipulated objects across \benchmarkname{}.

\section{Experiments}
\label{sec:experiments}

This section evaluates \modelname on 3D point trajectory prediction (Sec.~\ref{subsec:motion_prediction}) and on two downstream transfer tasks: robotic planning (Sec.~\ref{subsec:robotics}) and trajectory-guided video generation (Sec.~\ref{subsec:video_gen}). Model ablations are presented in Appendix~\ref{appsec:ablations}.

\subsection{3D point motion forecasting}
\label{subsec:motion_prediction}
We first evaluate \modelname and existing motion prediction methods on the 3D point trajectory prediction task in \benchmarkname{}.

\begin{table*}[t]
\centering
\small
\setlength{\tabcolsep}{4pt}
\renewcommand{\arraystretch}{1.12}
\resizebox{\textwidth}{!}{%
\begin{tabular}{@{}ll cc ccc ccc ccc@{}}
\multirow{2}{*}{\textbf{Paradigm}} & \multirow{2}{*}{\textbf{Model}} &
\multicolumn{2}{c}{\textbf{Inputs}} &
\multicolumn{3}{c}{\textbf{HOT3D~\cite{hot3d}}} & \multicolumn{3}{c}{\textbf{WorldTrack~\cite{feng2025st4rtrack}}} & \multicolumn{3}{c}{\textbf{DAVIS~\cite{davis}}} \\
\cmidrule(lr){3-4} \cmidrule(lr){5-7} \cmidrule(lr){8-10} \cmidrule(lr){11-13}
& & Frames & Text & ADE$\downarrow$ & FDE$\downarrow$ & PWT$\uparrow$
& ADE$\downarrow$ & FDE$\downarrow$ & PWT$\uparrow$
& ADE$\downarrow$ & FDE$\downarrow$ & PWT$\uparrow$ \\
\midrule
\multirow{2}{*}{\shortstack{Non- \\ parametric}}
& Static          & 1 & \xmark & $0.180$ & $0.316$ & $0.293$ & $0.167$ & $0.317$ & $0.390$ & $2.281$ & $4.360$ & $0.085$ \\
& Extrapolate     & 3 & \xmark & $0.159$ & $0.309$ & $0.351$ & $0.184$ & $0.432$ & $0.436$ & $2.683$ & $5.741$ & $0.104$ \\
\midrule
\multirow{2}{*}{Pixel-space}
& Wan2.2-5B~~\cite{wan22}       & 1 & \cmark & $0.200$ & $0.308$ & $0.253$ & $0.852$ & $1.046$ & $0.090$ & $3.074$ & $5.192$ & $0.051$ \\
& Cosmos Predict~\cite{nvidia2025cosmos}  & 5 & \cmark & $0.225$ & $0.294$ & $0.199$ & $0.831$ & $0.988$ & $0.072$ & $4.191$ & $6.368$ & $0.033$ \\
\midrule
\multirow{3}{*}{3D model}
& ObjectForesight~\cite{soraki2026objectforesightpredictingfuture3d} & 3 & \xmark & $\underline{0.129}$ & $\underline{0.192}$ & $0.353$ & -- & -- & -- & -- & -- & -- \\
& EgoScaler~\cite{Yoshida_2025_CVPR}       & 1 & \cmark & $0.170$ & $\mathbf{0.179}$ & $0.218$ & -- & -- & -- & -- & -- & -- \\
& Robot4DGen~\cite{liu2025geometryaware4dvideo}      & 3 & \xmark & $0.212$ & $0.271$ & $0.112$ & $0.548$ & $0.704$ & $0.121$ & $2.120$ & $3.382$ & $0.081$ \\
\midrule
2D track
& Track2Act~\cite{track2act}       & 1 & \xmark & $0.294$ & $0.413$ & $0.202$ & $1.230$ & $1.567$ & $0.053$ & $4.853$ & $8.110$ & $0.018$ \\
\midrule
\multirow{4}{*}{\shortstack{\textbf{3D track} \\ \textbf{(Ours)}}}
& \textbf{\modelname-FM} & 1 & \cmark & $0.183$ & $0.311$ & $0.286$ & $0.165$ & $0.305$ & $0.401$ & $1.380$ & $2.205$ & $\underline{0.165}$ \\
& \textbf{\modelname-FM} & 3 & \cmark & $0.135$ & $0.255$ & $\underline{0.382}$ & $0.158$ & $0.295$ & $\underline{0.438}$ & $1.480$ & $2.520$ & $0.130$ \\
& \textbf{\modelname-AR} & 1 & \cmark & $0.157$ & $0.290$ & $0.303$ & $\underline{0.148}$ & $\underline{0.269}$ & $0.424$ & $\mathbf{1.146}$ & $\mathbf{1.843}$ & $\mathbf{0.199}$ \\
& \textbf{\modelname-AR} & 3 & \cmark & $\mathbf{0.109}$ & $0.217$ & $\mathbf{0.444}$ & $\mathbf{0.143}$ & $\mathbf{0.261}$ & $\mathbf{0.445}$ & $\underline{1.227}$ & $\underline{2.108}$ & $0.153$ \\
\end{tabular}%
}
\caption{\textbf{3D point trajectory prediction on \benchmarkname{}.}
We report 3D ADE, FDE, and average PWT in meters. Input columns indicate the number of past observations consumed and whether the model accepts a natural-language action description. \textit{\modelname-AR} denotes our autoregressive variant, while \textit{\modelname-FM} denotes the flow-matching variant. 
}
\label{tab:motion_prediction}
\end{table*}
\begin{figure}[t]
  \centering
  \includegraphics[width=\linewidth]{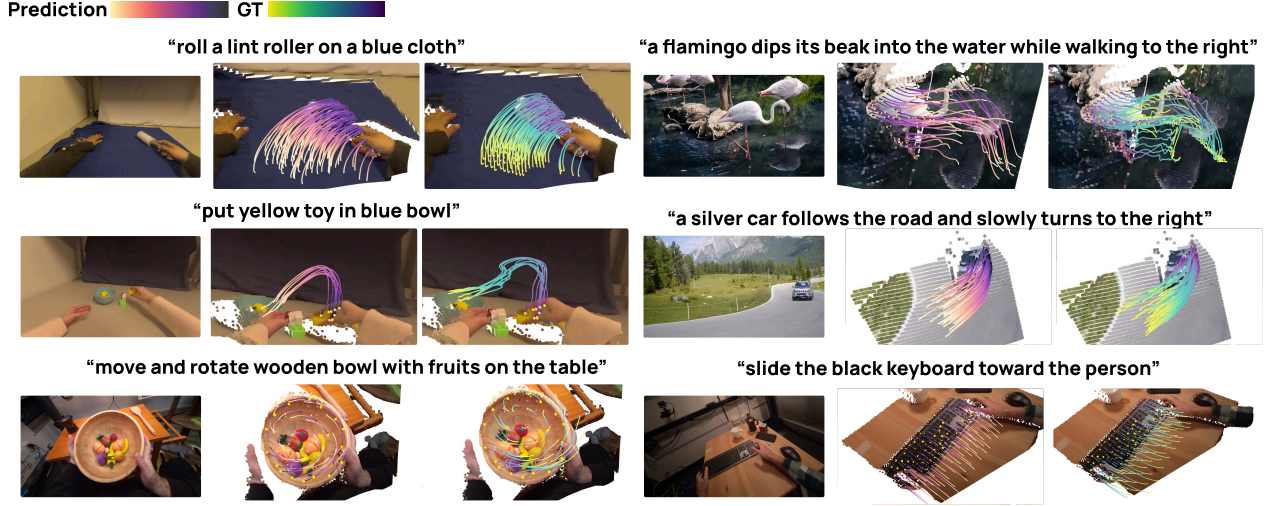}
  \caption{\textbf{Qualitative \modelname prediction.} \modelname predicts accurate motion trajectories on diverse motion patterns with different action instructions.
  }
  \label{fig:task1_prediction}
\end{figure}

\noindent\textbf{\modelname implementation.}
\modelname uses the pretrained 4B Molmo2~\cite{clark2026molmo2} as its VLM backbone. Training proceeds in two stages. In the first stage, we randomly sample a start timestamp $t_0$ from each video clip and sample $N=8$ query points from one object. The model is supervised to predict $T=8$ future timestamps at $15$ fps, giving $64$ future 3D point targets per example. 
We train with history length $H = 3$ in the first stage, providing a short visual history before $t_0$. This stage runs for $40$K steps. In the second stage, we continue training for $10$K steps while increasing the prediction horizon to $T=32$. In this stage we train two varients of models, with $H = 3$ and $H = 1$ respectively, to support both short visual history and single frame history settings.

\noindent \textbf{Baselines.} 
Baselines fall into four families. \textbf{Non-parametric baselines} include \textit{Static}, which keeps each point fixed at its initial 3D position, and \textit{Extrapolate}, which estimates a constant velocity from the history frames and linearly extrapolates into the future. \textbf{Pixel-space methods} first generate future RGB video and then recover query-point trajectories using the 3D tracking pipeline from Sec.~\ref{subsec:motion_annotation}; we compare against Wan2.2-I2V-5B~\cite{wan22} and Cosmos-Predict-2.5~\cite{nvidia2025cosmos}. \textbf{Parametric 3D prediction methods} predict an intermediate 3D representation; we include methods whose output can be converted to 3D point motion. ObjectForesight~\cite{soraki2026objectforesightpredictingfuture3d} and EgoScaler~\cite{Yoshida_2025_CVPR} forecast future 6-DoF object poses under a rigid-object assumption, and we use their predicted pose sequence to transform all query points. Robot4DGen predicts future scene flow~\cite{liu2025geometryaware4dvideo}, from which we extract query point trajectories. \textbf{2D point-track methods} predict future image-plane trajectories; we evaluate Track2Act by lifting its predicted 2D tracks to 3D using ground-truth depth~\cite{track2act}. Note that PointWorld~\cite{pointworld} and MotionForcast~\cite{thakkar2026forecastingmotion} are also closely related works, but their models have not yet been released.

\noindent \textbf{Evaluation setting and metrics.}
We evaluate all methods in \benchmarkname{}. Each method predicts future motion for up to $2$ seconds at $15$ fps; if the ground truth clip has a shorter valid future horizon, we evaluate only on the available frames. Since baselines vary in the number of input frames $H$ and whether they accept text condition, we follow each baseline's native setting where applicable. 
Following prior work~\cite{thakkar2026forecastingmotion}, we use best-of-$5$ evaluation for each sample.
We follow the point-track forecasting metrics used in~\cite{thakkar2026forecastingmotion}, adapted from 2D pixel distance to 3D metric distance. In our setting, one unit corresponds to one meter in 3D world coordinates. $\mathrm{ADE}$ measures the mean displacement error across all visible query points and predicted timesteps, while $\mathrm{FDE}$ measures the final-timestep displacement error. $\mathrm{PWT}$ is the average fraction of predicted points within $\{0.01,0.02,0.05,0.10,0.20\}$ meters of the ground truth.

\noindent\textbf{Results.} 
Tab.~\ref{tab:motion_prediction} shows the quantitative results. Note that ObjectForesight~\cite{soraki2026objectforesightpredictingfuture3d} and EgoScaler~\cite{Yoshida_2025_CVPR} are evaluated only on the HOT3D subset because they require object mesh inputs, which are available only for HOT3D. \modelname outperforms prior methods by a large margin in almost all subsets of \benchmarkname{}, with the autoregressive variant achieving the strongest overall performance.
The autoregressive model performs better than the flow-matching model under deterministic trajectory metrics, likely because conditioning on the previously generated coordinate sequence encourages temporally smooth predictions. Using three RGB observation frames generally improves over the single-frame setting, as additional history provides velocity cues for future motion. A notable finding is that simple non-parametric baselines are competitive with, or stronger than, several learned baselines, including pixel-space video prediction methods, suggesting that visually plausible RGB futures do not necessarily recover accurate metric point motion. 
We also show qualitative examples in Fig.~\ref{fig:task1_prediction}, where \modelname accurately predicts 3D motion trajectories across diverse motion patterns and language instructions.

\subsection{\modelname transfers effectively to robotics planning}
\label{subsec:robotics}
\begin{figure}[t]
  \centering
  \begin{subfigure}[b]{0.59\textwidth}
    \includegraphics[width=\linewidth]{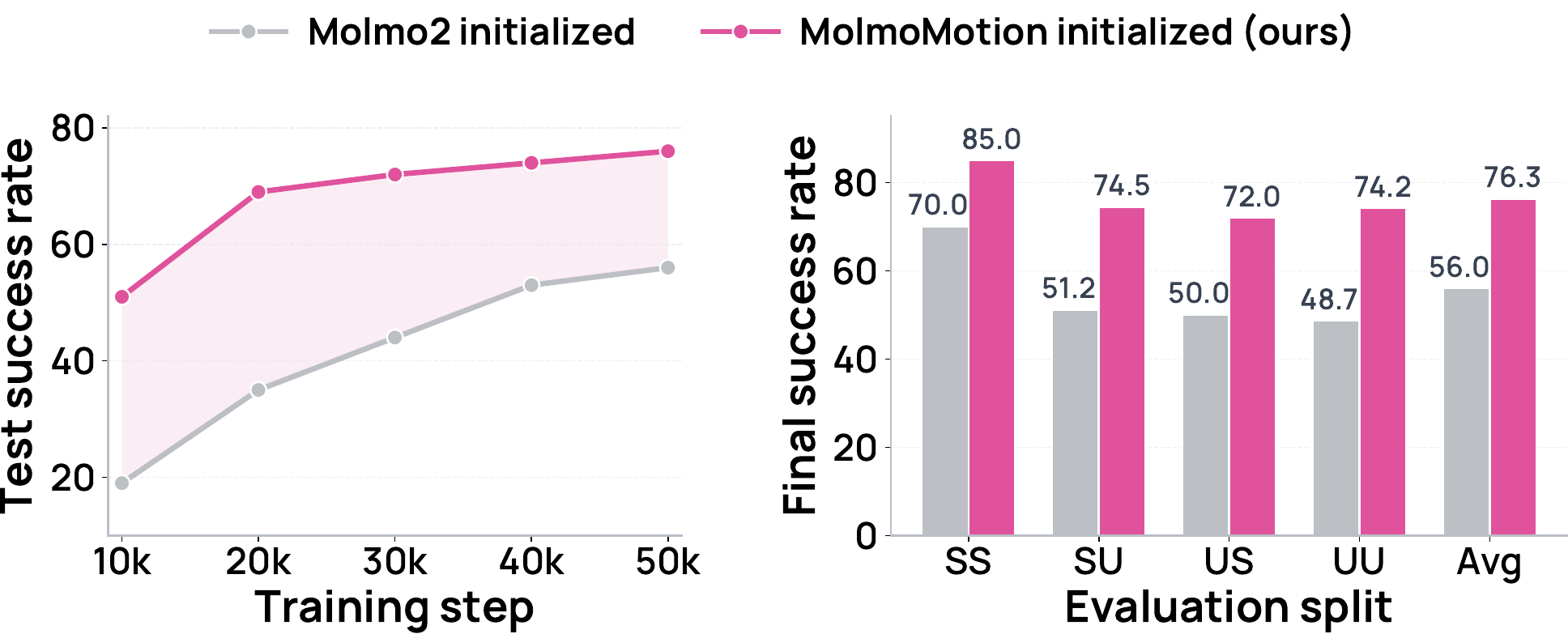}
    \caption{Pick-and-place task on the MolmoSpaces benchmark.}
    \label{fig:robot_pickplace}
  \end{subfigure}\hfill
  \begin{subfigure}[b]{0.40\textwidth}
    \includegraphics[width=\linewidth]{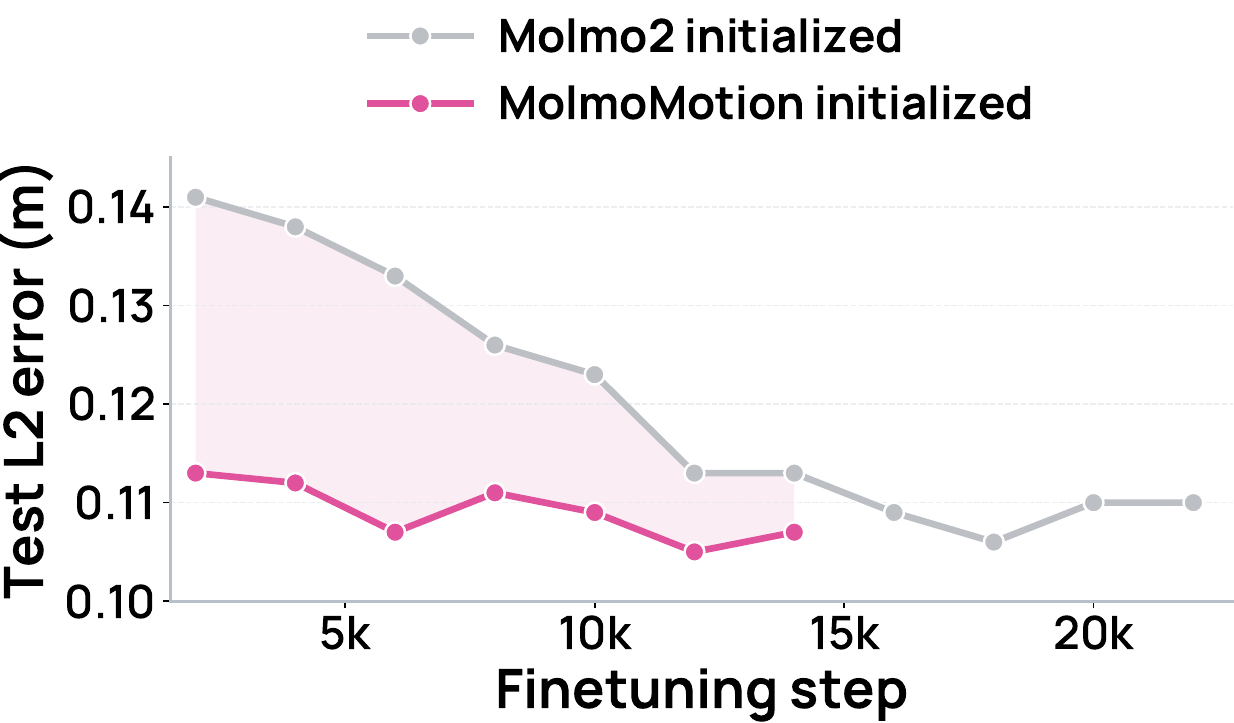}
    \caption{Trajectory finetuning on DROID.}
    \label{fig:robot_droid}
  \end{subfigure}
  \caption{\textbf{\modelname transfers effectively to robotics planning.} (a) Task success rate on the MolmoSpaces Franka Pick-and-place benchmark over robot-finetuning steps (left) and the final step per split breakdown (right) across SS (seen scene, seen object), SU (seen scene, unseen object), US (unseen scene, seen object), and All. (b) Test L2 error of predicted 3D trajectories on held-out DROID clips. In both settings, initializing from \modelname substantially outperforms the baseline.
  }
  \label{fig:robot_transfer}
\end{figure}

\begin{figure}[t]
  \centering
  \includegraphics[width=0.97\linewidth]{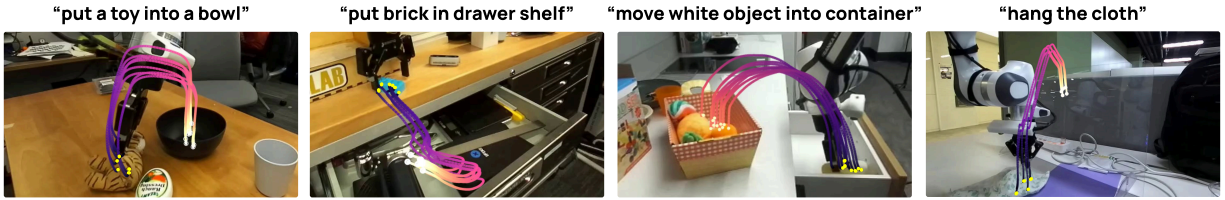}
  \caption{\textbf{\modelname trajectory prediction on real-world scenarios.} Predicted future 3D object trajectories on a held-out scenario after finetuning \modelname on DROID videos. \modelname can plan accurate object trajectories for various robotic manipulation tasks.}
  \label{fig:droid_prediction}
\end{figure}

The representation learned by \modelname is well-suited for robotics planning. The intuition is that object motion in 3D is more transferable than embodiment-specific actions. A human hand and a robot gripper execute differently, but successful manipulation often produces similar object trajectories in 3D space.

We use the pick-and-place task as a controlled transfer setting, focusing on the post-grasp stage where the policy must lift, transport and place the object at the correct target.
 We train two MolmoBot policies~\cite{deshpande2026molmobot} with the same flow-matching action head and 20K released episodes, differing only in backbone initialization: Molmo2 pretrained weights vs. \modelname-AR. During evaluation, the original MolmoBot policy performs grasping, after which control is handed to our trained policy. On FrankaPickandPlaceDroidMiniBench in MolmoSpaces~\cite{molmospaces}, we report closed-loop success across seen/unseen scene and object splits. As shown in Fig.~\ref{fig:robot_transfer}a, \modelname initialization substantially improves training efficiency and final performance: success reaches 51\% at 10K steps vs. 19\% for Molmo2, and final average success increases from 56.0\% to 76.3\%. The smaller drop in unseen-object and unseen-scene splits further suggests \modelname improves downstream generalization.

 We further evaluate whether the \modelname can be adapted to real-world robot scenarios. We finetune \modelname on DROID single-camera videos~\cite{droid} using the same 3D object point trajectory prediction task. Fig.~\ref{fig:droid_prediction} shows qualitative examples of object trajectory planned by \modelname across different scenes, objects, and tasks in DROID held out videos. We also  compare against training the same architecture with Molmo2 initialization. As shown in Fig. \ref{fig:robot_transfer}b, \modelname starts with substantially lower trajectory error and reaches the best performance much quicker. This suggests that the motion prior learned by \modelname can be efficiently adapted to real-world robot data. We leave closed-loop real-robot evaluation to future work.

\begin{figure}[t]
    \centering
    \includegraphics[width=1\linewidth]{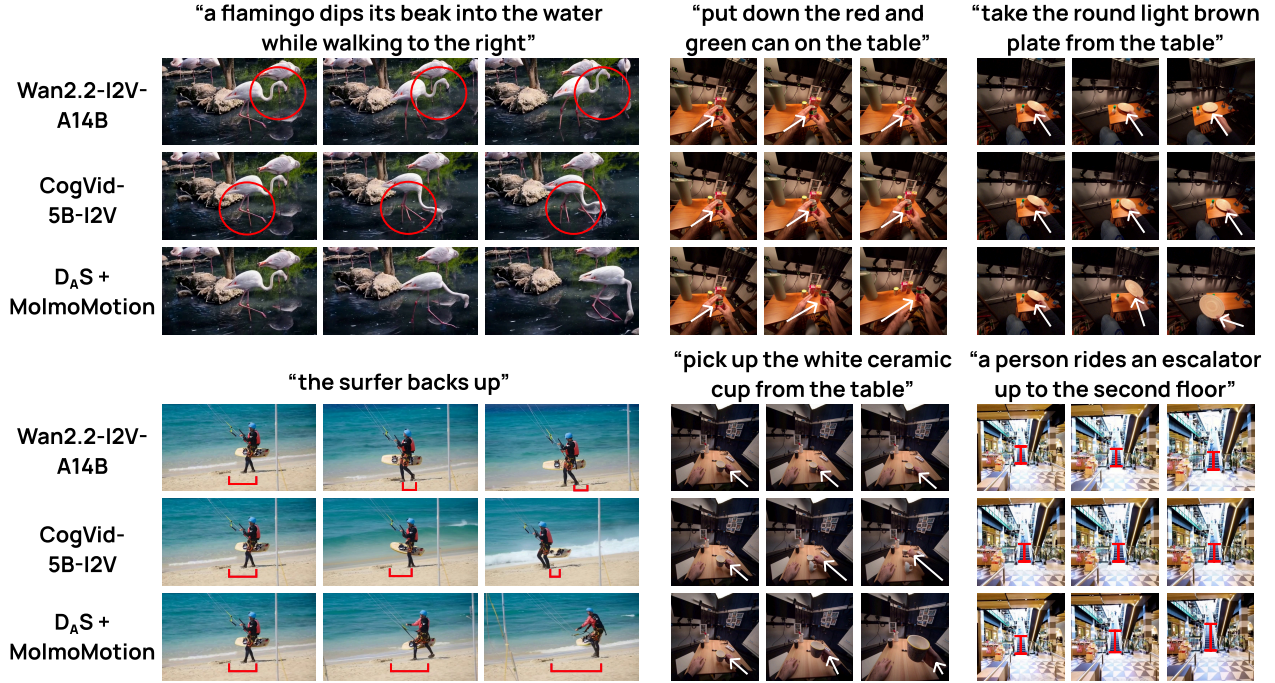}
    \caption{\textbf{Qualitative Video Generation Comparison.} \modelname-guided videos exhibit more physically plausible object motion and follow the prompted action more faithfully.}
    \label{fig:video_gen}
\end{figure}

\begin{table}[t]
  \centering
  \small
  \setlength{\tabcolsep}{6pt}
  \begin{tabular}{lccccc}
    Method & Tem-Con & Subj-Cons & M-Smooth & Dyn-Deg & Bg-Cons \\
    \midrule
    CogVideoX-5B~\cite{yang2025cogvideoxtexttovideodiffusionmodels}                 & $0.964$ & $0.939$ & $\underline{0.988}$ & $0.861$ & $0.941$ \\
    Wan-14B~\cite{wan22}                      & $\underline{0.965}$ & $\underline{0.940}$ & $0.983$ & $\mathbf{0.908}$ & $\underline{0.947}$ \\
    \textsc{DaS~\cite{gu2025diffusionasshader}}\,+\,MolmoMotion & $\mathbf{0.968}$ & $\mathbf{0.950}$ & $\mathbf{0.990}$ & $\underline{0.876}$ & $\mathbf{0.948}$ \\
  \end{tabular}
  \vspace{1mm}
\caption{\textbf{Video generation quality on \benchmarkname{} videos.} We report metrics in VBench~\cite{huang2023vbench} that evaluates motion: temporal consistency, subject consistency, motion smoothness, dynamic degree, and background consistency.
  }
  \label{tab:vid_gen_comparison}
\end{table}

\subsection{\modelname provides controllable motion for video generation}
\label{subsec:video_gen}

A natural question is whether the 3D point trajectories predicted by \modelname can serve as an explicit motion-control signal for downstream video generation. Given an input image and action description, we first obtain query points on the manipulated object and estimate their initial 3D positions using the pipeline in Sec.~\ref{subsec:motion_annotation}. \modelname then predicts their future 3D trajectories, which we use to guide DaS~\cite{gu2025diffusionasshader}, a 3D point-trajectory-guided image-to-video model built on CogVideoX-5B~\cite{yang2025cogvideoxtexttovideodiffusionmodels}. We compare against caption-conditioned image-to-video generators without explicit motion guidance, including DaS's base model CogVideoX-5B and the larger Wan2.2-I2V-A14B~\cite{wan22}. As shown in Tab.~\ref{tab:vid_gen_comparison}, DaS+\modelname improves over CogVideoX-5B on all metrics and outperforms Wan2.2-I2V-A14B on four out of five metrics. Qualitatively (Fig.~\ref{fig:video_gen}), it produces more physically plausible object motion, especially for fine-grained manipulation. It also follows the prompted action more faithfully. These results suggest that \modelname predicts 3D trajectories that can serve as an effective control interface for downstream video generation.

\section{Related work}
\label{sec:related_work}

\noindent\textbf{Motion prediction models.}
Motion prediction models differ mainly in the representation they forecast. Pixel-space methods cast prediction as conditional video generation, from latent-action world models~\cite{micheli2023iris, hu2023gaia1} to recent large-scale video generators~\cite{bruce2024genie, nvidia2025cosmos, wan22}; these models are general but spend substantial capacity rendering appearance, lighting, and camera motion. Alongside multimodal representation learning~\cite{zheng2023exif,zheng2024iterated}, latent prediction methods forecast learned feature states~\cite{zhou2025dinowm, assran2025vjepa2}, avoiding pixels but producing encoder-tied representations that are difficult to use as physical quantities. Recent point-trajectory forecasting methods predict category-agnostic motion directly~\cite{track2act, thakkar2026forecastingmotion}, but operate in 2D image space, where object motion is entangled with camera motion and viewpoint change. 3D forecasting methods predict physically grounded states such as human motion~\cite{mao2020learningtrajectorydependencieshuman}, rigid-object pose~\cite{Yoshida_2025_CVPR, soraki2026objectforesightpredictingfuture3d}, or scene-level 3D motion~\cite{pointworld}, but are often tied to specific object categories or domains. We predict object-attached 3D point trajectories in world coordinates, yielding a category-agnostic and physically grounded representation.

\noindent\textbf{3D point tracking.}
Our data annotation pipeline builds on recent advances in 3D point tracking from monocular video. 2D point trackers estimate temporally persistent pixel locations for queried points~\cite{doersch2022tapvid, doersch2023tapir}, with recent models improving long-range tracking~\cite{doersch2024bootstap}, occlusion reasoning~\cite{cotracker3}, and dense correspondence over entire videos~\cite{alltracker}. 
Recent monocular reconstruction methods lift image observations into 3D by estimating camera motion and per-frame depth, enabling pixel tracks to be lifted to 3D ~\cite{vipe, megasam}. 
There are also direct 3D point trackers that track points in 3D space~\cite{spatialtrackerv2, feng2025st4rtrack, zhang2025efficientlyreconstructingdynamicscenes}. 
Together, these methods have made it increasingly feasible to recover 3D motion from ordinary RGB videos, though the focus is on tracking observed frames rather than forecasting future motion.

\noindent\textbf{Human-to-robot transfer.}
Prior work transfers non-robot video to robot control through different abstractions. Some methods re-target human hands, endpoints, or skill trajectories as planning scaffolds~\cite{bahl2023affordances, vidbot,fang2026molmoact2, traj2action, bharadhwaj2024gen2act, li2025h2r}. Others train generalist vision-language-action policies on cross-embodiment robot data~\cite{bousmalis2023robocat, padalkar2024openx, octo2024, openvla, pi0}, sometimes with human video, but still predict embodiment-specific actions. Closest to ours are methods that learn motion priors or controls from unlabeled video through latent actions or inverse dynamics~\cite{bahl2023swim, dexwm, latentactionwild, lapa, egovla}. In contrast, we use 3D world-frame object point trajectories as the pretraining target itself, yielding a motion prior that is both embodiment-agnostic and directly usable for downstream robot learning.

\noindent\textbf{Trajectory representations and recent world-action models.}
TrajViT~\cite{zheng2025trajvit} groups observed video content into panoptic sub-object trajectory tokens, while TrajTok~\cite{zheng2026trajtok} learns trajectory tokenization with a differentiable segmenter and encoder for video understanding. These methods use trajectories to organize visual representations; \modelname instead predicts future object-attached 3D point trajectories in a shared world frame.
3DPWM~\cite{peri2026pointworldmodels} completes partial point clouds before learning action-conditioned dynamics, emphasizing completed scene geometry for long-horizon planning. PointZero~\cite{duisterhof2026pointzero} learns transferable 3D dynamics by completing point tracks from an RGB-D observation and sparse partial trajectories, with synthetic pretraining followed by adaptation to dynamics prediction and imitation learning.
EgoWAM~\cite{li2026egowam} studies human--robot co-training with alternative world-prediction targets, finding that DINO features and 3D motion flow transfer more effectively than pixels. ModAR~\cite{hung2026modar} sequentially predicts future point tracks, visual features, and depth before predicting robot actions. These approaches complement our focus on learning a category-agnostic 3D motion prior from large-scale video.

\section{Limitations} 
\modelname requires multiple forward passes to predict dense point tracks on an object,  since pretraining uses only 8 query points per example due to Molmo2 context length limit in stage 2 training. This sparse point setting is not sufficient to densely represent object geometry, limiting the model's understanding of fine-grained structure and complex deformable motion. In addition, more downstream evaluations are needed to fully establish the effectiveness of this motion pretraining task, such as closed-loop real-world robot experiments.

\section{Conclusions}
We introduce \modelname, a language-guided motion predictor that forecasts future trajectories of object-attached points in a 3D world frame. We built MolmoMotion-1M for pretraining and \benchmarkname{} for 3D motion evaluation. Experiments show that \modelname significantly outperforms prior motion prediction methods and transfers to robot manipulation and video generation.

\section*{Acknowledgements}

This work would not be possible without the support of our colleagues at Ai2. We thank David Albright, Stephen Kelman, Kristin Cha,  Byron Bischoff, David Everhart, Jon Borchardt, Kyle Wiggers, Will Smith, Peter Clark, Dieter Fox, and Noah Smith for their important work for the \modelname public release. We thank Ropedia for providing access to the Xperience dataset used in this work, and granting permission the release of MolmoMotion under the Apache License 2.0. Chenhao Zheng is partially funded through an Apple grant. We thank Oncel Tuzel, Pavan Kumar, and Rick Chang for the helpful discussion and support on this project.

\clearpage
\bibliographystyle{abbrvnat}
\bibliography{references}

\clearpage
\appendix
\section*{Appendix}
\begingroup
\hypersetup{hidelinks}
{\setlength{\parindent}{0pt}
\setlength{\parskip}{2pt}

\hyperref[appsec:qualitative]{\textbf{\ref*{appsec:qualitative}\quad Qualitative examples}}\dotfill\pageref{appsec:qualitative}

\hyperref[appsec:data_pipeline]{\textbf{\ref*{appsec:data_pipeline}\quad MolmoMotion-1M Data Generation Details}}\dotfill\pageref{appsec:data_pipeline}

\hspace*{2em}\hyperref[appsec:data_sources]{\ref*{appsec:data_sources}\quad Source video corpora}\dotfill\pageref{appsec:data_sources}

\hspace*{2em}\hyperref[appsec:phrase_extraction]{\ref*{appsec:phrase_extraction}\quad Recaptioning and object name extraction}\dotfill\pageref{appsec:phrase_extraction}

\hspace*{2em}\hyperref[appsec:grounding_lifting]{\ref*{appsec:grounding_lifting}\quad Semantic grounding and 3D lifting}\dotfill\pageref{appsec:grounding_lifting}

\hspace*{2em}\hyperref[appsec:filtering]{\ref*{appsec:filtering}\quad Trajectory filtering and smoothing}\dotfill\pageref{appsec:filtering}

\hyperref[subsec:appendix_3dworldbench]{\textbf{\ref*{subsec:appendix_3dworldbench}\quad \benchmarkname{}}}\dotfill\pageref{subsec:appendix_3dworldbench}

\hspace*{2em}\hyperref[subsec:appendix_hot3d]{\ref*{subsec:appendix_hot3d}\quad HOT3D}\dotfill\pageref{subsec:appendix_hot3d}

\hspace*{2em}\hyperref[subsec:appendix_worldtrack_pipeline]{\ref*{subsec:appendix_worldtrack_pipeline}\quad WorldTrack}\dotfill\pageref{subsec:appendix_worldtrack_pipeline}

\hspace*{2em}\hyperref[subsec:appendix_davis]{\ref*{subsec:appendix_davis}\quad DAVIS}\dotfill\pageref{subsec:appendix_davis}

\hspace*{2em}\hyperref[subsec:appendix_eval]{\ref*{subsec:appendix_eval}\quad Evaluation Protocol}\dotfill\pageref{subsec:appendix_eval}

\hyperref[appsec:model_details]{\textbf{\ref*{appsec:model_details}\quad Model Implementation Details}}\dotfill\pageref{appsec:model_details}

\hspace*{2em}\hyperref[appsec:model_arch]{\ref*{appsec:model_arch}\quad Architecture}\dotfill\pageref{appsec:model_arch}

\hspace*{2em}\hyperref[appsec:prompt_format]{\ref*{appsec:prompt_format}\quad Prompt format}\dotfill\pageref{appsec:prompt_format}

\hspace*{2em}\hyperref[appsec:training]{\ref*{appsec:training}\quad Training hyperparameters}\dotfill\pageref{appsec:training}

\hspace*{2em}\hyperref[appsec:flowmatching]{\ref*{appsec:flowmatching}\quad Flow-matching objective and inference}\dotfill\pageref{appsec:flowmatching}

\hyperref[appsec:ablations]{\textbf{\ref*{appsec:ablations}\quad Model Ablations}}\dotfill\pageref{appsec:ablations}

\hyperref[appsec:robotics]{\textbf{\ref*{appsec:robotics}\quad Robotics Transfer Settings}}\dotfill\pageref{appsec:robotics}

\hspace*{2em}\hyperref[appsec:molmospaces]{\ref*{appsec:molmospaces}\quad MolmoSpaces pick-and-place}\dotfill\pageref{appsec:molmospaces}

\hspace*{2em}\hyperref[appsec:droid]{\ref*{appsec:droid}\quad DROID trajectory finetuning}\dotfill\pageref{appsec:droid}

\hyperref[appsec:vidgen]{\textbf{\ref*{appsec:vidgen}\quad Video generation experiment details}}\dotfill\pageref{appsec:vidgen}

\hspace*{2em}\hyperref[appsec:vidgen_methods]{\ref*{appsec:vidgen_methods}\quad Methods compared}\dotfill\pageref{appsec:vidgen_methods}

\hspace*{2em}\hyperref[appsec:vidgen_eval]{\ref*{appsec:vidgen_eval}\quad Evaluation protocol}\dotfill\pageref{appsec:vidgen_eval}

}
\endgroup

\newpage

\section{Qualitative examples}
\label{appsec:qualitative}

\begin{figure}[p]
  \centering
  \includegraphics[width=\linewidth,keepaspectratio]{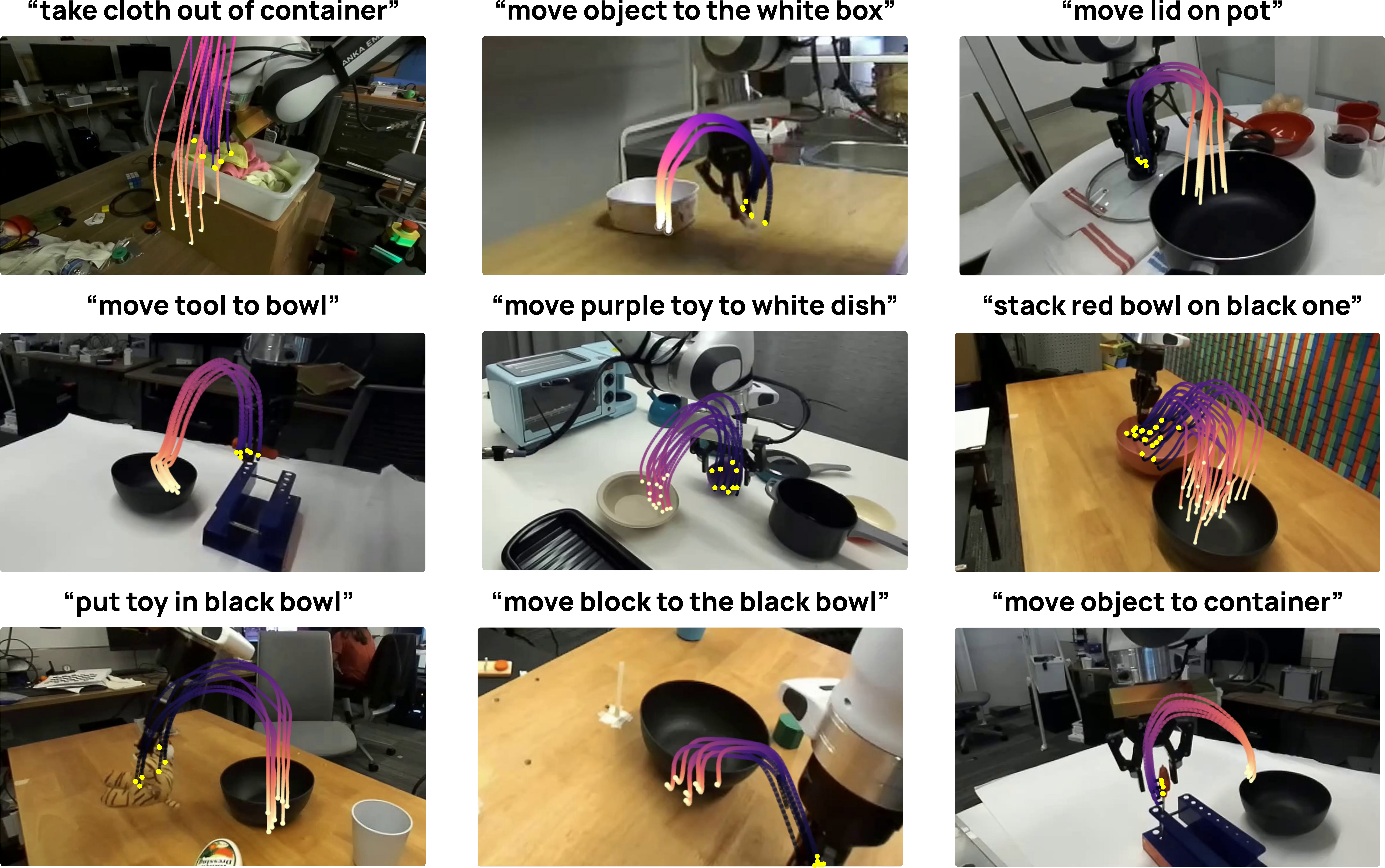}
  \caption{\textbf{\modelname predictions on held-out DROID clips.}}
  \label{fig:droid_appendix_qualitative}
\end{figure}

\begin{figure}[p]
  \centering
  \includegraphics[width=\linewidth,keepaspectratio]{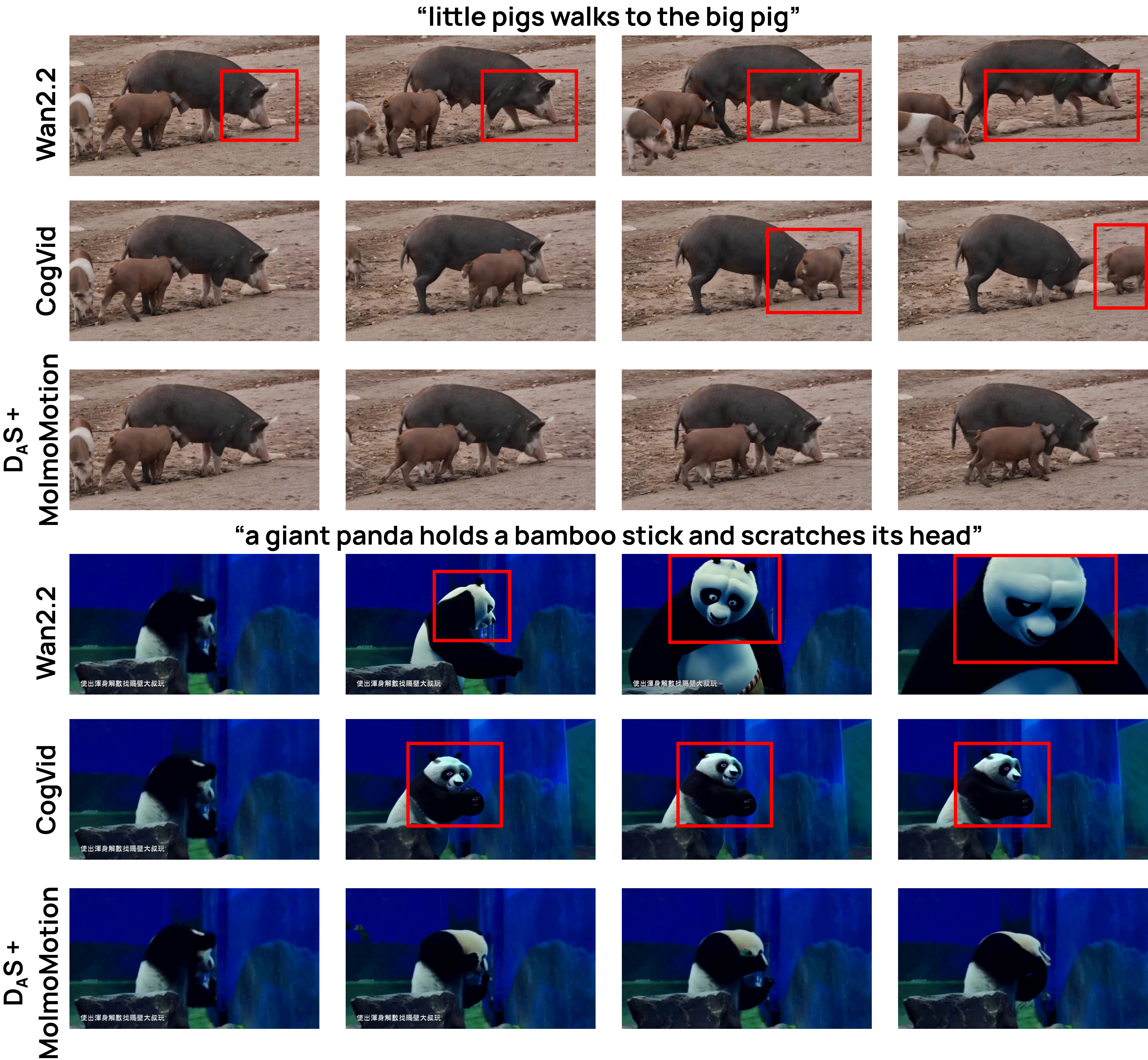}
  \caption{\textbf{Video generation comparisons on held-out \benchmarkname{} prompts (1/2).}}
  \label{fig:video_gen_appendix_qualitative_0}
\end{figure}

\begin{figure}[p]
  \centering
  \includegraphics[width=\linewidth,height=0.94\textheight,keepaspectratio]{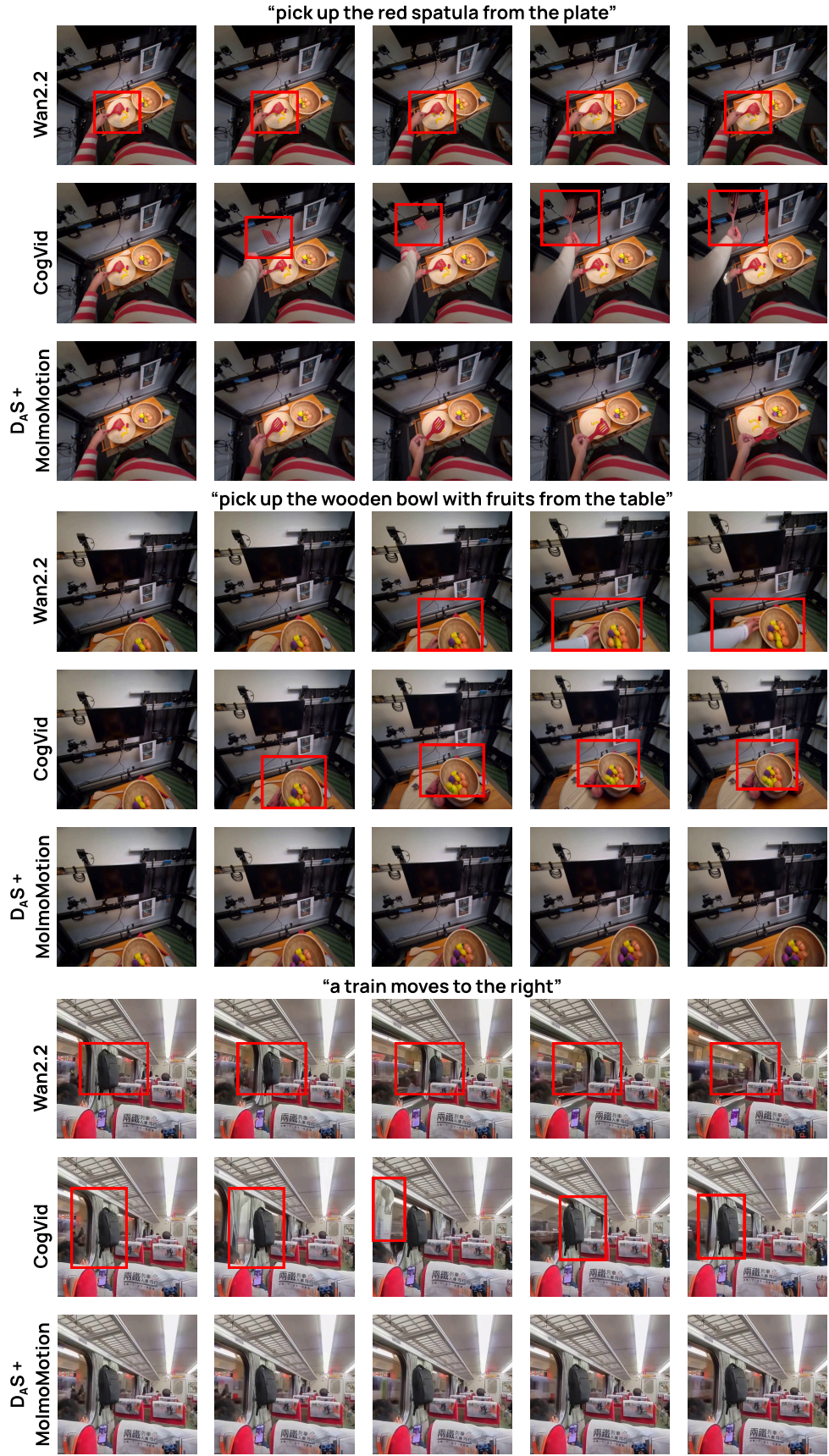}
  \caption{\textbf{Video generation comparisons on held-out \benchmarkname{} prompts (2/2).}}
  \label{fig:video_gen_appendix_qualitative_1}
\end{figure}

We provide additional qualitative results for two downstream applications of \modelname: real-robot trajectory prediction on DROID, and trajectory-conditioned video generation.

\noindent\textbf{Real-robot trajectory prediction on DROID.}
Fig.~\ref{fig:droid_appendix_qualitative} shows \modelname's predicted future 3D trajectories on held-out DROID clips after finetuning on real-robot video. \modelname plans accurate point trajectories across diverse manipulation scenes, objects, and tasks.

\noindent\textbf{Trajectory-conditioned video generation.}
Fig.~\ref{fig:video_gen_appendix_qualitative_0} and Fig.~\ref{fig:video_gen_appendix_qualitative_1} compare videos generated by DaS~\cite{gu2025diffusionasshader} conditioned on \modelname-predicted 3D trajectories with the unconditioned CogVideoX-5B and Wan-14B baselines on held-out \benchmarkname{} prompts. \modelname-guided videos exhibit more physically plausible object motion, preserve manipulated-object identity better, and follow the prompted action more faithfully than the unconditioned baselines.

Together these examples illustrate that \modelname{}'s 3D trajectories transfer to real-world robot data and serve as an effective control signal for downstream video generation, complementing the quantitative results in the main paper.

\section{MolmoMotion-1M Data Generation Details}
\label{appsec:data_pipeline}

This appendix expands the annotation pipeline summarized in Sec.~\ref{subsec:motion_annotation}. Given a public video and its caption, the pipeline recovers an object phrase that names the moving object, grounds the phrase to a set of query points on the object, and lifts those points into a metric 3D world frame. We describe the source corpora (\S\ref{appsec:data_sources}), recaptioning and object-name extraction (\S\ref{appsec:phrase_extraction}), object grounding and 3D lifting (\S\ref{appsec:grounding_lifting}), and trajectory filtering and smoothing (\S\ref{appsec:filtering}).

\subsection{Source video corpora}
\label{appsec:data_sources}

MolmoMotion-1M aggregates seven source corpora (Tab.~\ref{tab:appendix-sources}) that together cover egocentric and third-person human manipulation, simulated and robot manipulation, and in-the-wild scenes. The five manipulation corpora ship with action descriptions or task templates, which we use directly. YT-VIS supplies object masks but no captions, so we re-caption each clip from the video. Stereo4D contributes short third-person clips with metric stereo depth, expanding coverage to outdoor scenes and deformable subjects.

\begin{table}[h]
  \centering
  \small
  \setlength{\tabcolsep}{4pt}
  \begin{tabular}{lrll}
    Corpus     & Motion clips            & Domain            & Action-description source \\
    \midrule
    EgoDex      & $\sim$160K              & egocentric, human & VLM re-caption \\
    HD-EPIC     & $\sim$21K               & egocentric, human & narrations \\
    Xperience   & $\sim$500K              & 3rd-person, human & metadata \\
    MolmoSpaces & $\sim$185K              & 3rd-person, sim   & task templates \\
    DROID       & $\sim$27K               & 3rd-person, robot & language instructions \\
    YT-VIS      & $\sim$2K                & 3rd-person, wild  & VLM caption \\
    Stereo4D    & $\sim$70K               & 3rd-person, wild  & paper captions \\
  \end{tabular}
  \vspace{2mm}
  \caption{Source corpora used to construct MolmoMotion-1M.}
  \label{tab:appendix-sources}
\end{table}

\subsection{Recaptioning and object name extraction}
\label{appsec:phrase_extraction}

For corpora that do not ship object-level captions (EgoDex, where the original task labels are too coarse to ground a specific entity, and YT-VIS), we generate a one-sentence visual description of each clip with Molmo2-8B~\cite{clark2026molmo2}. The full 15\,FPS re-encoded video is passed to Molmo2-8B. The prompt is:
\begin{verbatim}
Watch this video carefully. Describe the manipulation action you
observe in exactly this format: [action verb] [specific object with
color/material/shape] [preposition and location if present].
Examples: "pick up red ceramic coffee mug", "place blue plastic
bottle on table". Be specific about the object -- include its color,
material, and shape as you see them.
Output only the short description, no extra words.
\end{verbatim}

Then we extract the manipulated object as a noun phrase with Qwen3-0.6B~\cite{yang2025qwen3}.

\subsection{Semantic grounding and 3D lifting}
\label{appsec:grounding_lifting}

Given an object phrase, we localize the entity, segment it, sample query points, track those points across the video, and lift the tracks to a metric 3D world frame.

\noindent\textbf{Localization with motion-aware prompting.}
We first localize the object as a 2D point with MolmoPoint-8B~\cite{clark2026molmopoint}, and then convert that point into a segmentation mask with SAM\,3~\cite{sam3}. The non-trivial choice here is the prompt given to MolmoPoint. Conditioning the prompt on the agent and the action, we give the model the following prompt: \textit{"point to the \{obj\} gripped and picked up by the hand"} for human-manipulation corpora, and \textit{"track the \{obj\}"} for in-the-wild video. This is substantially more reliable than a bare \textit{"point to the \{obj\}"} prompt. The motion cue disambiguates vague phrases like \textit{"an object on the table"} by biasing MolmoPoint toward the moving entity rather than a static distractor that matches the phrase equally well.

\noindent\textbf{Point prompt for SAM\,3.}
We feed the 2D point returned by MolmoPoint to SAM\,3. We then sample $N=100$ query points per mask using $K$-means cluster centers on the mask pixel coordinates, so points are spread across the object surface rather than concentrated near the centroid.

\noindent\textbf{2D tracking and lifting.}
We propagate query points through the video with AllTracker~\cite{alltracker}, which yields temporally persistent 2D tracks and per-frame visibility scores. We run ViPE~\cite{vipe} on the same video to estimate per-frame metric depth, intrinsics, and camera-to-world poses in a single pass. Each visible 2D track location is back-projected with the estimated depth and intrinsics, then transformed by the corresponding camera pose into the world frame anchored at the query-time camera.

\subsection{Trajectory filtering and smoothing}
\label{appsec:filtering}

Lifted 3D trajectories are corrupted by noise. We make this prior precise as follows.

\noindent\textbf{Anchor tracks and inconsistency score.}
We select a small set of anchor tracks $\mathcal{A}$ per object (sixteen in our pipeline) that are visible throughout the clip and have low per-frame velocity, taking these as the most reliable estimate of the object's body motion. For every other track $n$ we measure how its distance to each anchor varies over time:
\begin{equation}
e_n(t) \;=\; \mathrm{median}_{k \in \mathcal{A}}\;\bigl|\, \|\tilde{\mathbf{p}}_t^{n} - \tilde{\mathbf{p}}_t^{k}\|_2 \;-\; \bar d_{nk} \,\bigr|,
\qquad
\bar d_{nk} \;=\; \mathrm{median}_t\, \|\tilde{\mathbf{p}}_t^{n} - \tilde{\mathbf{p}}_t^{k}\|_2 .
\end{equation}
A rigid or near-rigid object keeps its inter-point distances roughly constant, so $e_n(t)$ is large precisely at frames where track $n$ has drifted in 2D, had its depth pulled to the background, or been displaced by a pose error.

\noindent\textbf{Trust weights and the choice of scale.}
We convert the inconsistency score into a per-frame trust weight $w_{t,n} = \exp\bigl(-e_n(t)/s_n\bigr)$ with a per-track scale $s_n$. We use $s_n = \mathrm{median}_t(e_n)$, which behaves well under uniform noise while still suppressing per-frame outliers within a track.

\noindent\textbf{Spatial auto-split.}
A single object phrase sometimes resolves to two physically separate instances, and mixing tracks across instances violates the rigid-object assumption behind $e_n(t)$. We therefore cluster points by their temporal-mean 3D position with mean-shift and run anchor selection, trust scoring, and smoothing independently per sub-cluster.

\noindent\textbf{Track-level outlier drop.}
Within each sub-cluster, we drop tracks whose mean trust $\bar w_n$ is a MAD-scaled z-score below the sub-cluster median.

\noindent\textbf{Depth-ray smoothing.}
Surviving tracks are still noisy along the depth axis. Following Stereo4D~\cite{stereo4d}, we re-parametrize the 3D point at frame $t$ as $\mathbf{p}_t^{n} = \mathbf{c}_t + \lambda_t^{n}\mathbf{r}_t^{n}$, where $\mathbf{c}_t$ is the camera center and $\mathbf{r}_t^{n}$ is the unit ray through the 2D track location, and we optimize the depth scalars $\{\lambda_t^{n}\}$ with two competing objectives:
\begin{equation}
\min_{\{\lambda_t^{n}\}}\;\;
\underbrace{\sum_{t,n} w_{t,n}^{\,2}\,\bigl\|\,\mathbf{c}_t + \lambda_t^{n}\mathbf{r}_t^{n} - \tilde{\mathbf{p}}_t^{n}\,\bigr\|_2^{2}}_{\text{trust-weighted pin to lifted point}}
\;\;+\;\;
\beta\;\underbrace{\sum_{n}\sum_{\Delta\in\{1,3,5\}}\sum_t \bigl\|\,\mathbf{p}_{t+\Delta}^{n} - 2\mathbf{p}_t^{n} + \mathbf{p}_{t-\Delta}^{n}\,\bigr\|_2^{2}}_{\text{multi-stride acceleration penalty}} .
\end{equation}
The pin term is gated by the trust weights, so untrusted frames are free to slide along the ray while trusted frames remain anchored; the acceleration term enforces smooth motion in 3D. Two design choices are worth explaining. First, the acceleration term is summed over multiple strides $\Delta\in\{1,3,5\}$ rather than only $\Delta=1$. A single-stride second difference penalizes only sharp single-frame jitter, but our depth errors include slow ramps over five to ten frames (e.g.\ when ViPE depth gradually bleeds onto a textureless background); penalizing acceleration at $\Delta=3$ and $\Delta=5$ in addition to $\Delta=1$ catches these slower drifts. Second, we optimize with first-order gradient descent rather than LBFGS. LBFGS converges faster on well-behaved tracks but diverges on near-degenerate tracks, so the more conservative optimizer is the safer default at corpus scale.

\section{\benchmarkname{}}
\label{subsec:appendix_3dworldbench}

\benchmarkname{} evaluates object-centric 3D motion forecasting across three
diverse datasets; Tab.~\ref{tab:appendix-3dworldbench-sources} summarizes the
per-source clip counts, resolutions, and frame rates. The following sections
detail the data preparation pipelines and evaluation protocol (including
metric definitions).

\begin{table}[h]
  \centering
  \small
  \setlength{\tabcolsep}{6pt}
  \begin{tabular}{lrlr}
    Dataset    & Clips & Resolution                & FPS \\
    \midrule
    HOT3D      & 497   & $1408\times1408$          & 30  \\
    WorldTrack & 155   & split-dependent$^\dagger$ & 30  \\
    DAVIS      & 90    & $854\times480$            & 24  \\
  \end{tabular}
  \vspace{2mm}
  \caption{Source datasets used to construct \benchmarkname{}.
    $^\dagger$\texttt{adt\_mini} $512\times512$; \texttt{ds\_mini} $1280\times720$;
    \texttt{po\_mini} $960\times540$; \texttt{pstudio\_mini} $640\times360$.}
  \label{tab:appendix-3dworldbench-sources}
\end{table}

\noindent\textbf{Exclusions.}
We exclude $10$ HOT3D clips that contain no ground-truth-visible query points at
frame~$0$ (AllTracker crashes at initialization with an empty point set).
For WorldTrack, we exclude the \texttt{tum} split entirely (no 3D tracks
shipped) and the $27$ \texttt{po\_mini} \texttt{dancingroom1\_3rd*}
sequences that exhibit cluttered motion patterns. No DAVIS sequences are
excluded. The same exclusion lists apply across all three baselines so that
comparisons are always over identical clip sets, leaving $497$ HOT3D, $155$
WorldTrack, and $90$ DAVIS clips.

\noindent\textbf{Human verification.}
For HOT3D and WorldTrack, we view each clip alongside the annotated
trajectories and confirm that (i) the foreground object is correctly
identified and (ii) the one-sentence model-generated description accurately reflects the
observed motion and action; clips that fail are re-annotated or excluded.
For DAVIS, we directly wrote the descriptions during trajectory review rather than reviewing model-generated captions.

\noindent\textbf{Contamination check.}
\benchmarkname{} contains no clip-level near-duplicates with the training
data of the evaluated models. All per-clip captions are either written directly by human
annotators or generated by Molmo2-8B and subsequently verified by annotators,
ensuring no captions are sourced from existing dataset metadata that evaluated models may have been trained on.

\subsection{HOT3D}
\label{subsec:appendix_hot3d}

HOT3D~\cite{hot3d} Aria provides $1{,}415$ source clips with per-frame 3D object pose
annotations as fitted mesh models. We sample $2{,}000$ surface points per
object at its first active frame and propagate them forward with per-frame
pose transforms, yielding world-space trajectories and their corresponding
pixel-space projections.

Each source clip typically contains one or two manipulated objects alongside
several static ones. We split every clip into single-moving-object sub-clips by detecting
per-object motion windows: for each object, we threshold a per-frame
body-speed signal (median surface-point displacement) at $0.005$\,m/frame
(${\approx}15$\,cm/s at $30$\,fps), bridge single-frame gaps, and drop runs
shorter than $0.5$\,s. This yields $2{,}534$ sub-clips, each
covering exactly one continuously moving foreground object. We then uniformly
subsample one sub-clip per group of five (random selection within the group),
yielding $507$ clips; the $10$ with no ground-truth-visible query points at frame~$0$
are excluded, leaving $497$.

\subsection{WorldTrack}
\label{subsec:appendix_worldtrack_pipeline}

WorldTrack~\cite{feng2025st4rtrack} bundles four indoor motion-capture splits
mixing egocentric and third-person viewpoints: \texttt{adt\_mini},
\texttt{ds\_mini}, \texttt{po\_mini}, and \texttt{pstudio\_mini} (resolutions
listed in Tab.~\ref{tab:appendix-3dworldbench-sources}). Unlike HOT3D,
source point tracks are not pre-assigned to objects. We augment the raw data
through a four-step pipeline: (1)~identify dynamic foreground points,
(2)~cluster them into per-object groups, (3)~segment sequences into
motion-coherent sub-clips, and (4)~generate a one-sentence caption per clip.

\begin{table}[h]
  \centering
  \small
  \setlength{\tabcolsep}{4pt}
  \resizebox{0.5\linewidth}{!}{%
  \begin{tabular}{llll}
    Dataset                & Filter type   & Window(s) & Threshold                                    \\
    \midrule
    \texttt{adt\_mini}     & per-frame     & 3, 5 fr   & 0.25\%             \\
    \texttt{po\_mini}      & global        & ---       & 1.0\%      \\
    \texttt{pstudio\_mini} & global+per-fr & 3, 15 fr  & 1.0\%         \\
    \texttt{ds\_mini}      & per-frame     & 10 fr     & 0.75\%             \\
  \end{tabular}%
  }
  \vspace{2mm}
  \caption{Per-dataset motion-filter settings for dynamic point extraction of WorldTrack data.}
  \label{tab:appendix-worldtrack-filter}
\end{table}

\noindent\textbf{Dynamic point extraction.}
We lift camera-space tracks into a shared world coordinate frame using
per-frame extrinsics; for \texttt{pstudio\_mini} (fixed camera), camera-space
coordinates serve directly as world-space. A point is classified as dynamic if
its world-space displacement over a look-back window exceeds a
scene-normalized threshold (a per-dataset fraction of the scene's
1st-to-99th percentile spatial extent; see
Tab.~\ref{tab:appendix-worldtrack-filter}). Masks from multiple window
lengths are OR-ed together, and each active segment is extended backward to
capture motion onset.

\noindent\textbf{Object clustering.}
We cluster the dynamic points into per-object groups in two stages.

\noindent\textbf{Stage~1 --- Temporal clustering (Leiden~\cite{traag2019leiden}).}
We build a pairwise affinity graph over the dynamic points. For each pair
$(i,j)$, let $\mathcal{T}_\text{co}$ be the set of co-visible frames and
$\mathbf{v}_i(t)$ the unit-normalized frame-to-frame velocity of point~$i$
at frame~$t$, with zero-velocity frames contributing~$0$. The affinity is
\[
  W(i,j) \;=\; \frac{1}{|\mathcal{T}_\text{co}|}
    \sum_{t\,\in\,\mathcal{T}_\text{co}}
    \langle\mathbf{v}_i(t),\,\mathbf{v}_j(t)\rangle.
\]
Pairs with no co-visible support or below a similarity threshold are zeroed
out. Additionally, any pair whose 3D world-space distance exceeds a fixed
fraction of scene extent at any sampled keyframe is forced apart---a single
violating frame is conclusive evidence of distinct objects, since rigid-body
points cannot drift far apart in 3D. We run Leiden community detection on the
resulting graph and iteratively merge undersized clusters into their most
temporally similar neighbor. Per-dataset minimum cluster sizes are $1$
(\texttt{adt\_mini}, \texttt{pstudio\_mini}), $3$ (\texttt{ds\_mini}), and
$5$ (\texttt{po\_mini}), set by visual inspection.

\noindent\textbf{Stage~2 --- Segmentation merging (SAM~2~\cite{sam2}).}
Temporal clustering occasionally splits a single physical object across
multiple clusters. For each cluster, we designate a representative point and
sample three keyframes evenly across its active window, then query SAM~2. Two
clusters are merged only when their representatives are spatially proximate
and co-occur within the same SAM~2 mask in at least one keyframe without ever
appearing in separate masks. The criterion is intentionally conservative: a
single frame of visible separation blocks the merge, since merging distinct
objects corrupts ground truth irreversibly.

\noindent\textbf{Sub-clip extraction and captioning.}
We group objects with overlapping active spans, bridge gaps of fewer than
three frames, split on longer gaps, and merge sub-clips shorter than $2$\,s
into their nearest neighbor. Each sub-clip is captioned by
Molmo2-8B~\cite{clark2026molmo2}, prompted to produce a one-sentence
egocentric description (e.g.\ \textit{``a hand moves a keyboard across the
desk''}); captions are subsequently verified by human annotators.

\subsection{DAVIS}
\label{subsec:appendix_davis}

DAVIS~\cite{davis} provides RGB frames and per-object segmentation masks but no 3D ground
truth. We generate 3D trajectories by running the MolmoMotion-1M annotation
pipeline, seeding query points from the DAVIS masks and lifting them through
ViPE depth and camera pose estimation. We verify the resulting tracks and write the per-clip action descriptions directly
during trajectory review.

\subsection{Evaluation Protocol}
\label{subsec:appendix_eval}

\noindent\textbf{Conditioning and future split.}
All metrics are computed on future frames only. With $T_\text{cond}\in\{1,3\}$,
the model observes frames $0,\ldots,T_\text{cond}-1$ and predicts frames
$T_\text{cond},\ldots,T-1$. Scoring is restricted to points visible at
frame~$0$, as these are the only query positions supplied to the model.
The evaluation mask is
\[
  v_\text{eval}[t,n] \;=\; v[t,n]\;\wedge\;v[0,n],
\]
applied before every metric. Let $\mathcal{S} = \{(t,n) : t \geq
T_\text{cond},\; v_\text{eval}[t,n]=1\}$ be the set of scored
(frame,\,point) pairs, and let $N_\text{eval} = \sum_n v[0,n]$ be the
per-clip count of evaluated points.

\noindent\textbf{Temporal alignment.}
HOT3D and WorldTrack ground truth tracks are at $30$\,fps, and DAVIS ground truth tracks are at 24 fps; Wan2.2 and Cosmos predictions are at $24$\,fps.
For HOT3D and Worldtrack, we resample these predictions to the ground truth
timebase by linearly interpolating positions at source time
$t_\text{src} = 0.8\,t_\text{gt}$ and nearest-neighbor-resampling
visibility; only the $\min(T_\text{gt},\,T_\text{resampled})$ overlapping
frames are scored. Track2Act outputs exactly $8$ frames at ground truth indices
$\mathrm{round}(i\,(T-1)/7)$ for $i=0,\ldots,7$; we match ground truth frames at the
same indices directly, with no interpolation.

\noindent\textbf{Metrics.}
All metrics are computed in 3D world space. Let $\hat{p}(t,n)$ and $q(t,n)$
denote the predicted and ground-truth positions of query point~$n$ at
frame~$t$ (in metres; $1$~unit~$=$~$1$\,m throughout).

\textbf{ADE} (Average Displacement Error):
\[
  \mathrm{ADE} \;=\; \frac{1}{|\mathcal{S}|}
    \sum_{(t,n)\,\in\,\mathcal{S}} \|\hat{p}(t,n) - q(t,n)\|_2
\]

\textbf{FDE} (Final Displacement Error), evaluated at the last frame $T-1$:
\[
  \mathrm{FDE} \;=\; \frac{1}{|\mathcal{S}_T|}
    \sum_{n\,\in\,\mathcal{S}_T} \|\hat{p}(T-1,n) - q(T-1,n)\|_2,
  \qquad \mathcal{S}_T = \{n : v_\text{eval}[T-1,n]=1\}.
\]

\textbf{PWT} (Percentage Within Threshold) at threshold $\delta$:
\[
  \mathrm{PWT}(\delta) \;=\;
    \frac{\bigl|\{(t,n)\in\mathcal{S} :
      \|\hat{p}(t,n)-q(t,n)\|_2 < \delta\}\bigr|}{|\mathcal{S}|}
\]
We report $\overline{\mathrm{PWT}}$, the mean of $\mathrm{PWT}(\delta)$ over
$\delta \in \{0.01, 0.02, 0.05, 0.10, 0.20\}$\,m. ADE and FDE are in meters
(lower is better); $\overline{\mathrm{PWT}}\in[0,1]$ (higher is better).

\section{Model Implementation Details}
\label{appsec:model_details}

We expand the architecture and training recipe here.

\subsection{Architecture}
\label{appsec:model_arch}

\noindent\textbf{Vision-language backbone}
We initialize from the public \texttt{Molmo2-4B-Pretrain} checkpoint. The
vision encoder is a SigLIP2 ViT operating on $378{\times}378$ RGB inputs at
14-pixel patch size, producing a $27{\times}27$ grid of 1152-D patch tokens
per frame. The Molmo2 connector pools the patch grid by $3{\times}3$ and
projects the pooled tokens through an MLP into the LLM hidden dimension of
$2560$. The language model is Qwen3-4B. All backbone parameters are trained
end-to-end.

\noindent\textbf{Decoder heads}
The autoregressive variant uses the unmodified Molmo2 LM head plus a small
regex parser that maps the answer span back to coordinates at inference. The
flow-matching variant uses a DiT trajectory expert with 36 blocks (one per LM
layer). Each block applies a self-attention over the trajectory-token tensor
of shape $(N, H{+}T, 3)$ followed by a cross-attention whose keys and values
come from the corresponding LM-layer hidden states and whose queries come from
the trajectory tokens. RoPE is applied along both the point-index and the
frame-index axes so the same self-attention can distinguish ``the same point
at different times'' from ``different points at the same time'' without
architectural specialization.

\subsection{Prompt format}
\label{appsec:prompt_format}

The autoregressive variant serializes everything as a single multimodal
prompt. Image tokens for the anchor frame and the $H$ history frames are
inserted by Molmo2's video preprocessor at the front; the textual portion
follows. With $N=8$ and $T=32$, an example prompt is

\begin{quote}\small\ttfamily
Predict the future 3D point coordinates of 8 points over 32 timestamps,\\
given action: ``open the drawer'',\\
2d history point features: ``<points anchor 1 <2d\_feat\_start><|point\_feat|><2d\_feat\_end>\\
\hphantom{2d history point features: ``<points anchor }2 <2d\_feat\_start><|point\_feat|><2d\_feat\_end>\\
\hphantom{2d history point features: ``<points anchor }\ldots{} 8 <2d\_feat\_start><|point\_feat|><2d\_feat\_end>/>'',\\
and history 3d point coordinates:\\
\hphantom{and }``<tracks coords=`0.0 1 0 0 0 \ 2 12 -3 0 \ldots\ 8 5 4 -1'>3d object history</tracks>''.
\end{quote}

The supervised answer span is

\begin{quote}\small\ttfamily
<tracks coords=``1.0 1 4 -8 2 \ 2 17 -10 5 \ldots\ 8 9 7 -3;\\
\hphantom{<tracks coords=``}2.0 1 8 -16 4 \ldots;\\
\hphantom{<tracks coords=``}\ldots\\
\hphantom{<tracks coords=``}32.0 1 \ldots\,''>3d object trajectories</tracks>
\end{quote}

Each per-frame block lists $(n,\ q_x,\ q_y,\ q_z)$ quadruples where $n$ is the
integer point identifier and $(q_x,q_y,q_z)$ are the millimetre-quantised
anchor-relative deltas $\bar{\boldsymbol{\delta}}_{t}^{n}$. Only points visible at frame $t$
are emitted; occluded points are not imputed. At inference, the model is asked
to generate this answer span autoregressively under greedy decoding, and the
parser re-assembles $\hat{\mathbf{P}}_{t_0+1:t_0+T}$ from the recovered
quadruples.

\subsection{Training hyperparameters}
\label{appsec:training}

We train with AdamW using the Molmo2 supervised-fine-tuning defaults
($\beta_1=0.9$, $\beta_2=0.95$, weight decay $0.1$). The learning rate is
warmed up linearly over the first $1$K steps to its peak and then cosine-decayed
to $0.1\times$ peak. Activations are bf16 with fp32 master weights; the model is
distributed with FSDP2 full-shard across $16$ H100 GPUs at per-device batch $16$
for a global batch size of $256$. Gradients are clipped at maximum-norm $1.0$.
Multi-dataset mixing across the six MolmoMotion-1M sources is square-root-weighted
by per-dataset clip count. The autoregressive cross-entropy is computed on the
answer span only, with the prompt portion (image, text, point-feature, and
history-coordinate tokens) masked out of the loss; the flow-matching MSE is
computed on the future positions only, leaving the clean history portion of
the trajectory tensor unsupervised.

\subsection{Flow-matching objective and inference}
\label{appsec:flowmatching}

We give the full flow-matching specification briefly described in
Sec.~\ref{subsec:model}. The flow-matching head predicts the future
anchor-relative trajectory tensor
$\boldsymbol{\delta}_{t_0+1:t_0+T} \in \mathbb{R}^{N \times T \times 3}$
in continuous metric coordinates, conditioned on the multimodal context
$\mathcal{C}$ (image, text, point-feature, and history-coordinate tokens) and
on the clean initial 3D query coordinates
$\{\boldsymbol{\delta}_{t_0}^{n}\}_{n=1}^{N}$.

\noindent\textbf{Forward interpolation.}
For each training example we draw a flow timestep $\tau \sim \mathcal{U}(0,1)$
and a noise tensor $\epsilon \in \mathbb{R}^{N \times T \times 3}$ whose
entries are i.i.d.\ standard Gaussian. The interpolated trajectory is
\begin{equation}
\boldsymbol{\delta}_{\tau}
\;=\;
(1-\tau)\,\epsilon
\;+\;
\tau\,\boldsymbol{\delta}_{t_0+1:t_0+T} ,
\label{eq:appendix-flow-interp}
\end{equation}
which slides linearly from pure Gaussian noise at $\tau=0$ to the clean
ground-truth future trajectory at $\tau=1$.

\noindent\textbf{Velocity head.}
The DiT decoder $v_\phi$ predicts the velocity field at $(\boldsymbol{\delta}_\tau, \tau)$
in the direction of the clean future. It takes three inputs:
(i) the noised trajectory tensor with clean history concatenated to the noisy
future, RoPE-tagged along both the point-index and the frame-index axes;
(ii) the scalar $\tau$ via a sinusoidal embedding added to every trajectory
token; and (iii) the per-layer Molmo2 LM hidden states $\mathcal{C}$
through the per-block cross-attention described in
\S\ref{appsec:model_arch}.

\noindent\textbf{Training loss.}
The decoder is trained with the standard
flow-matching mean-squared error~\cite{lipman2023flowmatching},
\begin{equation}
\mathcal{L}_{\mathrm{FM}}
\;=\;
\mathbb{E}_{\tau,\,\epsilon}
\!\left[\,
\left\|
v_{\phi}\!\left(
\boldsymbol{\delta}_{\tau},\,\tau,\,
\{\boldsymbol{\delta}_{t_0}^{n}\}_{n=1}^{N},\,
\mathcal{C}
\right)
-
\bigl(
\boldsymbol{\delta}_{t_0+1:t_0+T} - \epsilon
\bigr)
\right\|_2^2
\,\right] ,
\label{eq:appendix-flow-loss}
\end{equation}
where the regression target $\boldsymbol{\delta}_{t_0+1:t_0+T} - \epsilon$
is the constant velocity that drives $\boldsymbol{\delta}_\tau$ along the
straight-line path of Eq.~\eqref{eq:appendix-flow-interp}. The loss is
masked to the future positions only, leaving the clean history portion of
the trajectory tensor unsupervised.

\noindent\textbf{Inference.}
At inference we sample $\epsilon \sim \mathcal{N}(0, I)$ and integrate the
learned velocity field with $K=10$ Euler steps, advancing $\tau$ uniformly
from $0$ to $1$ in increments of $\Delta\tau = 0.1$. Each step evaluates
$v_\phi$ once and updates
\begin{equation}
\boldsymbol{\delta}_{\tau + \Delta\tau}
\;=\;
\boldsymbol{\delta}_{\tau}
\;+\;
\Delta\tau \cdot
v_{\phi}\!\left(
\boldsymbol{\delta}_{\tau},\,\tau,\,
\{\boldsymbol{\delta}_{t_0}^{n}\}_{n=1}^{N},\,
\mathcal{C}
\right) .
\label{eq:appendix-euler}
\end{equation}
The final $\boldsymbol{\delta}_1$ is added to the anchor 3D position
$\mathbf{p}_{\mathrm{anc}}$ to recover the predicted future positions in
the world frame, $\hat{\mathbf{P}}_{t_0+1:t_0+T}$. Because the DiT block
count matches the LM layer count and one $v_\phi$ evaluation reuses cached
LM activations, each Euler step has roughly the cost of one LM forward
pass.

\noindent\textbf{Comparison to autoregressive decoding.}
We report flow-matching numbers in Tab.~\ref{tab:motion_prediction}
alongside the autoregressive variant. The autoregressive head is stronger on
the deterministic point-prediction metrics ADE / FDE / PWT because each
predicted coordinate is conditioned in a strict left-to-right sense on every
previously emitted coordinate, which encourages temporally smooth rollouts.
The flow-matching head, by contrast, samples from the conditional
distribution rather than committing to a single mode, which is desirable in
settings where the action description leaves multiple plausible futures.

\begin{table}[t]
\centering
\small
\setlength{\tabcolsep}{4pt}
\renewcommand{\arraystretch}{1.12}
\resizebox{\linewidth}{!}{%
\begin{tabular}{@{}l ccc ccc ccc@{}}
\multirow{2}{*}{\textbf{Variant}} &
\multicolumn{3}{c}{\textbf{HOT3D}} &
\multicolumn{3}{c}{\textbf{WorldTrack}} &
\multicolumn{3}{c}{\textbf{DAVIS}} \\
\cmidrule(lr){2-4} \cmidrule(lr){5-7} \cmidrule(lr){8-10}
& ADE$\downarrow$ & FDE$\downarrow$ & PWT$\uparrow$
& ADE$\downarrow$ & FDE$\downarrow$ & PWT$\uparrow$
& ADE$\downarrow$ & FDE$\downarrow$ & PWT$\uparrow$ \\
\midrule
\textbf{MolmoMotion-AR ($H=3$)} &
$\mathbf{0.109}$ & $\mathbf{0.217}$ & $\mathbf{0.444}$ &
$\mathbf{0.143}$ & $\mathbf{0.261}$ & $\mathbf{0.445}$ &
$\mathbf{1.227}$ & $\mathbf{2.108}$ & $\mathbf{0.153}$ \\
\midrule
without 2D point feature &
$0.118$ & $0.231$ & $0.421$ &
$0.155$ & $0.282$ & $0.418$ &
$1.310$ & $2.252$ & $0.143$ \\
Absolute coords (no delta) &
$0.165$ & $0.330$ & $0.276$ &
$0.220$ & $0.401$ & $0.288$ &
$1.940$ & $3.275$ & $0.082$ \\
without language instruction &
$0.158$ & $0.318$ & $0.291$ &
$0.215$ & $0.392$ & $0.305$ &
$1.890$ & $3.182$ & $0.092$ \\
\midrule
$N=16$ query points &
$0.106$ & $0.212$ & $0.452$ &
$0.140$ & $0.255$ & $0.451$ &
$1.198$ & $2.061$ & $0.158$ \\
\end{tabular}%
}
\vspace{5pt}
\caption{\textbf{Model ablations on \benchmarkname{}.} All variants use the
autoregressive head with the same Molmo2-4B initialization, the same
$40$K$+10$K training schedule, and the same evaluation protocol as
Tab.~\ref{tab:motion_prediction}. The reference row is MolmoMotion-AR at
$H=3$. Rows 2--4 remove one design choice; row 5 doubles the per-object query
count. Anchor-relative parameterization and language conditioning are the
largest contributors; the 2D point feature yields a small but consistent
gain; doubling $N$ improves accuracy slightly at substantial training cost
(see text).}
\label{tab:ablations}
\end{table}

\section{Model Ablations}
\label{appsec:ablations}

We ablate three load-bearing design choices in the autoregressive
variant: the per-query-point 2D feature, the anchor-relative coordinate
parameterization, and the language-instruction conditioning.
All ablations use the autoregressive variant with everything else held
identical to the main model.

\noindent\textbf{Ablations.}
The four ablated variants are constructed as follows.
(i) \textit{Without 2D point feature.} The grid-sampled per-query-point feature
is omitted to test if 2D point feature is useful.
(ii) \textit{Absolute coordinates.} We replace the anchor-relative deltas
$\boldsymbol{\delta}_t^n$ with absolute world-frame positions $\mathbf{p}_t^n$
in both the prompt's history block and the answer span. The $1$ mm
quantization grid is preserved; only the coordinate origin changes. This
isolates the value of the anchor-relative parameterization introduced in
Sec.~\ref{subsec:model}.
(iii) \textit{Without language instruction.} The action caption $a$ is
replaced with a single fixed token (``\texttt{motion}''); image, text,
point-feature, and history-coordinate tokens are unchanged. This isolates the
value of language conditioning.
(iv) \textit{$N=16$ query points.} We double the number of query points sampled
per object from $N=8$ to $N=16$, leaving every other hyperparameter unchanged.
This tests whether denser per-object coverage improves prediction accuracy and
quantifies the cost of going beyond our default.

\noindent\textbf{Results.}
Tab.~\ref{tab:ablations} reports 3D ADE, FDE, and PWT on the three
\benchmarkname{} splits. Removing the anchor-relative parameterization produces
the largest single drop ($\approx 50\%$ on ADE/FDE across all splits), making
absolute coordinates the strongest signal of design importance. Removing the
language instruction produces a comparable drop, indicating that the action
description does substantially more than disambiguate the object: it provides
the direction prior the model relies on when the visual context alone leaves
the future ambiguous, with the largest hit on DAVIS where intent is hardest to
infer from a single anchor frame. The 2D point feature contributes a smaller
but consistent gain ($5$--$8\%$ on ADE/FDE, $\approx 5\%$ on PWT) and is most
helpful on DAVIS, where the manipulated object is small relative to the frame.
Doubling the per-object query count from $N=8$ to $N=16$ improves accuracy by
roughly $2$--$3\%$ on ADE/FDE and $\approx 2\%$ on PWT across all three
splits. The gain is small because the eight default points already cover the
manipulated object's surface densely after $K$-means selection, and our query
points are sampled to be spatially well-distributed. The cost, however, is
substantial: with $T=32$ in stage~2, the autoregressive answer span is twice
as long under $N=16$ and exceeds the $4096$-token context window of the
Qwen3-4B language model that backs Molmo2-4B. We therefore keep $N=8$ as the
default, which trades a small accuracy improvement for the longer prediction
horizon $T=32$. Lifting this constraint --- through tokenization schemes that
encode multiple coordinates per LM token, or through context-extension
recipes for the backbone --- is a natural next step for representing dense
object coverage at long horizons; we leave it to future work.

\noindent\textbf{Inference cost.}
The two heads share a single Molmo2-4B forward pass over the prompt and
diverge only in how they decode the future trajectory. AR emits the
answer span as text, so its cost grows linearly in $N{\cdot}T$. FM runs
a fixed $K$ Euler steps through the DiT trajectory expert, so its cost
is independent of $T$. Tab.~\ref{tab:latency} reports per-clip latency
at the headline $(N{=}8, T{=}32)$ setting on a single A100. Flow-matching
is roughly two orders of magnitude faster than autoregressive decoding,
which is the regime that matters in closed-loop robotic control,
trajectory-conditioned video generation, and large-scale evaluation.

\begin{table}[h]
\centering
\small
\setlength{\tabcolsep}{6pt}
\renewcommand{\arraystretch}{1.12}
\begin{tabular}{@{}l c c@{}}
Variant & Setting & Latency (s / clip) \\
\midrule
\modelname-AR & $N{=}8,\ T{=}8$  & $37.2$ \\
\modelname-AR & $N{=}8,\ T{=}32$ & $148.4$ \\
\modelname-FM ($K{=}10$) & $N{=}8,\ T{=}32$ & $\mathbf{1.1}$ \\
\end{tabular}
\vspace{5pt}
\caption{\textbf{Inference cost.} Per-clip latency at $H{=}3$ on a single
A100. AR cost scales linearly in $N{\cdot}T$; FM cost is independent of
$T$. Flow-matching is roughly $\mathbf{150\times}$ faster than
autoregressive decoding at $T{=}32$, at the modest accuracy cost in
Tab.~\ref{tab:motion_prediction}.}
\label{tab:latency}
\end{table}

\section{Robotics Transfer Settings}
\label{appsec:robotics}

This appendix documents the implementation specifics of the two robotics
experiments reported in Sec.~\ref{subsec:robotics}: closed-loop pick-and-place
on MolmoSpaces (Fig.~\ref{fig:robot_transfer}a) and 3D-trajectory finetuning
on DROID (Fig.~\ref{fig:robot_transfer}b).

\subsection{MolmoSpaces pick-and-place}
\label{appsec:molmospaces}

\noindent\textbf{Policy.}
The downstream policy follows the MolmoBot~\cite{deshpande2026molmobot} prompt-encoder
recipe. The Molmo2-4B vision-language backbone is followed by an ActionExpert
head: a flow-matching transformer with $36$ cross-attention blocks (one per
LM layer), action dimension $8$ (7 arm joints plus 1 gripper), and action
horizon $16$. At inference, action chunks are produced by integrating the
ActionExpert's velocity field for $10$ Euler steps from Gaussian noise. The
backbone is initialised either from \modelname (autoregressive variant after
both training stages) or from the public \texttt{Molmo2-4B-Pretrain}
checkpoint for the matched control.

\noindent\textbf{Inputs.}
Each step provides three history frames per camera at sim-step deltas
$\{-4, -2, 0\}$ from two cameras (egocentric exo and wrist-mounted), plus the
current 8-D robot state normalised to unit-quantile space using statistics
fit on the training split. The textual prompt reuses the
trajectory-prediction wrapper of \S\ref{appsec:prompt_format} with an empty
answer span; the LM head's cross-entropy weight is set to zero, so only the
ActionExpert flow-matching MSE contributes to the gradient.

\noindent\textbf{Action representation.}
Actions are absolute joint-position targets supervised over a 16-step chunk
($\approx 1.06$\,s of simulated time at $15$\,fps). At execution time we run
$8$ steps of the predicted chunk and re-query the policy at sim-step $+8$,
matching the training-time action-time semantics. Per-joint deltas are
clamped to $\pm 0.2$\,rad/step at execution as a safety guard against
out-of-distribution velocity predictions.

\noindent\textbf{Training.}
We finetune on the $20$K pick-and-place episodes released with MolmoBot,
restricted to \texttt{policy\_phase} $\in \{5, 6, 7, 8\}$ (lift, preplace,
place, retreat) so the policy is supervised only on the post-grasp portion of
each trajectory. Optimization uses AdamW with the Molmo2 SFT defaults,
maximum sequence length $2048$, gradient clipping at norm $1.0$, bf16
mixed precision, FSDP2 full-shard, per-device batch $2$ with $8$
gradient-accumulation micro-batches for a global batch size of $128$ across
$8$ H100 GPUs. Training runs for $100$K optimization steps; checkpoints are
saved every $5$K steps and evaluated periodically.

\noindent\textbf{Hybrid rollout and evaluation.}
Closed-loop evaluation uses a hybrid rollout. The released
\texttt{MolmoBot-DROID} policy drives every episode through the approach and
grasp phases. The first sim step at which the simulator reports
\texttt{held = true} on the pickup object marks a hand-off trigger; one
execute-horizon window ($8$ sim steps) later, control is handed to the
evaluated policy, which performs the lift, transport, and placement. Both
policies output absolute joint-position targets. Each rollout runs for at most
$600$ sim steps ($\approx 40$\,s at $15$\,fps); an episode succeeds when the
simulator's terminal \texttt{success} flag is true, i.e.\ when the pickup
object lies inside the success-position threshold of its target receptacle.

\subsection{DROID trajectory finetuning}
\label{appsec:droid}

\noindent\textbf{Task and data.}
We use the same 3D point-trajectory prediction objective as in
Sec.~\ref{subsec:model}; no robot-action loss is applied. DROID wrist and
shoulder videos are paired with 3D trajectories produced by running the
MolmoMotion-1M annotation pipeline of Sec.~\ref{subsec:motion_annotation} on
the DROID corpus, yielding the same triplet of (object mask, action
description, dense 3D trajectory) as the pretraining data. We hold out a fixed 10 percent subset of clips for the test set.

\noindent\textbf{Initialization and finetuning.}
The pretraining-init run starts from the MolmoMotion-AR ($H=1$) checkpoint; the matched control starts
from \texttt{Molmo2-4B-Pretrain} and is trained on DROID alone. Both runs use
the same hyperparameters:
AdamW, $128$ global batch on $8$ H100s, max-norm $1.0$ gradient clipping,
bf16 mixed precision. Both runs are trained for $12$K finetuning steps; we
evaluate every $1$K steps and trace the test-loss curves shown in
Fig.~\ref{fig:robot_transfer}b.

\noindent\textbf{Evaluation.}
We report 3D test L2 on held-out DROID clips, computed identically to the
ADE metric of Sec.~\ref{subsec:motion_prediction} but on the DROID test
split. The pretraining-init run starts at substantially lower L2 and reaches
the matched-control's $12$K-step error after only $\approx 2$K finetuning
steps.

\begin{table}[t]
  \centering
  \small
  \setlength{\tabcolsep}{6pt}
  \begin{tabular}{ll}
    Metric              & What it measures \\
    \midrule
    Tem-Con$\uparrow$   & mean CLIP cosine similarity between adjacent frames \\
    Subj-Cons$\uparrow$ & feature-similarity of the moving subject across frames \\
    M-Smooth$\uparrow$  & smoothness of the estimated optical-flow residual \\
    Dyn-Deg$\uparrow$   & fraction of clips whose flow magnitude exceeds a motion threshold \\
    Bg-Cons$\uparrow$   & feature-similarity of the background region across frames \\
  \end{tabular}
  \vspace{2mm}
  \caption{Higher-is-better video-quality metrics used to compare the three image-to-video generators. Subj-Cons, M-Smooth, Dyn-Deg, and Bg-Cons are standard VBench dimensions; Tem-Con is computed separately from CLIP frame embeddings.}
  \label{tab:vidgen-metrics}
\end{table}

\section{Video generation experiment details}
\label{appsec:vidgen}

We expand the video-generation experiment summarized in Sec.~\ref{subsec:video_gen}: starting from the first frame and action description of a benchmark clip, we ask three image-to-video generators to synthesize the rest of the clip and compare their outputs on standard video-quality metrics. We describe the methods compared (\S\ref{appsec:vidgen_methods}) and the evaluation protocol (\S\ref{appsec:vidgen_eval}).

\subsection{Methods compared}
\label{appsec:vidgen_methods}

We evaluate three image-to-video generators. \textsc{DaS}~\cite{gu2025diffusionasshader} is a 3D-track-conditioned image-to-video model built on the CogVideoX-5B backbone~\cite{yang2025cogvideoxtexttovideodiffusionmodels}, and is the only method that consumes \modelname's predicted 3D trajectories. CogVideoX-5B-I2V is the same backbone without the tracking branch; comparing \textsc{DaS}+\modelname against it isolates the contribution of track conditioning while controlling for the underlying generator. Wan2.2-I2V-A14B~\cite{wan22} is an unconditioned image-to-video baseline at roughly $2.8\times$ the parameter count of CogVideoX-5B-I2V, and tells us how a much larger generator without explicit motion conditioning compares to a smaller one with it.

\subsection{Evaluation protocol}
\label{appsec:vidgen_eval}

For each clip we feed the same first-frame image and caption to all three generators; \textsc{DaS} additionally receives the MolmoMotion prediction track. Each generated clip is then transformed to the ground-truth frame count, frame rate, and resolution, so all methods are scored on the same temporal and spatial grid.

We score each generated clip on five higher-is-better video-quality metrics (Tab.~\ref{tab:vidgen-metrics}): four standard VBench dimensions~\cite{huang2023vbench}---subject consistency, motion smoothness, dynamic degree, and background consistency---and a CLIP-based temporal-consistency score, reported alongside as a complementary check on frame-to-frame coherence. We aggregate over \benchmarkname{} (\S\ref{subsec:appendix_3dworldbench}).

\end{document}